\documentclass[11pt]{article}

\usepackage[preprint]{acl}

\usepackage{times}
\usepackage{latexsym}

\usepackage[T1]{fontenc}

\usepackage[utf8]{inputenc}

\usepackage{microtype}

\usepackage{inconsolata}

\usepackage{graphicx}
\usepackage{booktabs}
\usepackage{multirow}
\usepackage{enumitem}

\usepackage{amsmath}
\usepackage{amssymb}

\usepackage{tcolorbox}
\usepackage{cuted}

\title{GSEM: Graph-based Self-Evolving Memory for Experience Augmented Clinical Reasoning}

\author{
  Xiao Han\textsuperscript{\ensuremath{\dagger}},
  Yuzheng Fan\textsuperscript{\ensuremath{\dagger}},
  Sendong Zhao\textsuperscript{*},
  Haochun Wang and Bing Qin \\
  Research Center for Social Computing and Interactive Robotics, \\
  Harbin Institute of Technology, China \\
  \texttt{\{xhan, yzfan, sdzhao\}@ir.hit.edu.cn}
}

\begin{document}
\maketitle

\begingroup
\renewcommand\thefootnote{\fnsymbol{footnote}}
\setcounter{footnote}{0}
\footnotetext[2]{Equal contribution.}
\footnotetext[1]{Corresponding author.}
\endgroup
\setcounter{footnote}{0}

\begin{abstract}
Clinical decision-making agents can benefit from reusing prior decision experience. However, many memory-augmented methods store experiences as independent records without explicit relational structure, which may introduce noisy retrieval, unreliable reuse, and in some cases even hurt performance compared to direct LLM inference. We propose \textbf{GSEM} (\textbf{G}raph-based \textbf{S}elf-\textbf{E}volving \textbf{M}emory), a clinical memory framework that organizes clinical experiences into a dual-layer memory graph, capturing both the decision structure within each experience and the relational dependencies across experiences, and supporting applicability-aware retrieval and online feedback-driven calibration of node quality and edge weights. Across MedR-Bench and MedAgentsBench with two LLM backbones, GSEM achieves the highest average accuracy among all baselines, reaching 70.90\% and 69.24\% with DeepSeek-V3.2 and Qwen3.5-35B, respectively. Code is available at \url{https://github.com/xhan1022/gsem}.
\end{abstract}

\section{Introduction}
\label{sec:intro}

LLM agents have demonstrated strong capabilities in medical decision-making, from diagnostic reasoning to multi-step problem solving \citep{medagents, MDAgents}. Yet real-world clinical deployment involves a continuous stream of high-stakes cases with incomplete information, strong condition dependence, and substantial uncertainty \citep{MediQ}. In such settings, relying solely on parametric knowledge can be insufficient for adapting to the diverse and condition-specific patterns that emerge across cases. This motivates an explicit experience memory that accumulates prior decision knowledge and supports future cases through targeted retrieval and reuse.

\begin{figure}[t]
    \centering
    \includegraphics[width=\columnwidth]{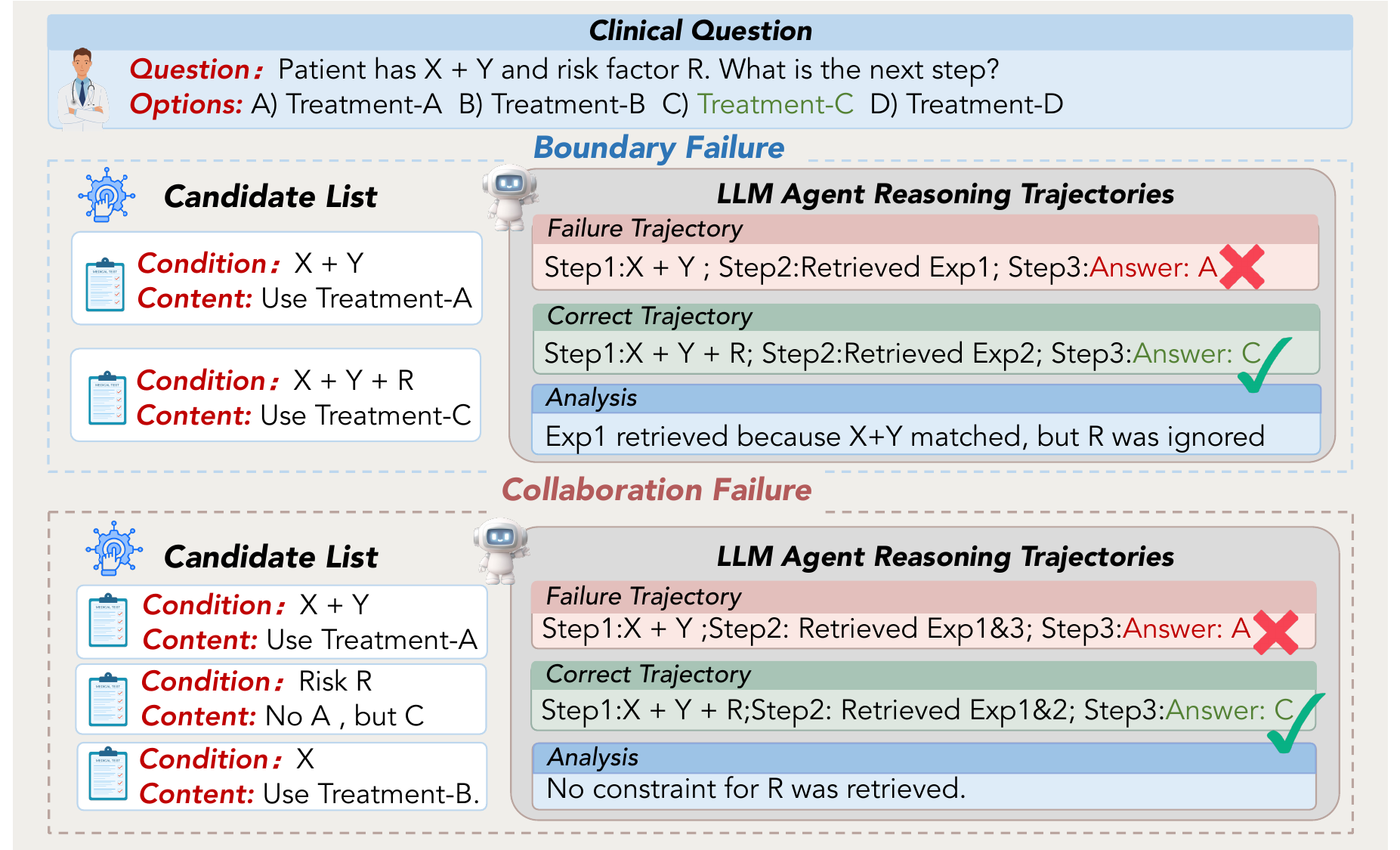}
    \caption{Two motivating failure cases of experience-augmented clinical reasoning. \textbf{Boundary Failure} (top): an experience is retrieved based on partial condition match, while critical boundary conditions are ignored. \textbf{Collaboration Failure} (bottom): conflicting experiences without joint-use value are co-retrieved, leading to incoherent guidance for the final decision. A detailed clinical case study is provided in Appendix~\ref{app:case_study}.}
    \label{fig:intro_motivation}
\end{figure}

Despite growing interest in agent memory, building experience memory that is both structurally organized and reliably applicable remains challenging. Figure~\ref{fig:intro_motivation} motivates two key challenges for effective experience memory: (i) boundary-aware applicability checking and (ii) relation-aware composition when multiple experiences are used together. Yet many retrieval-augmented pipelines treat applicability implicitly and rely primarily on similarity-based top-$k$ retrieval, leaving applicability boundaries hard to verify \citep{selfrag, rarr}. Meanwhile, memory-augmented methods usually store experiences as independent entries \citep{xu2025mem} without explicit representations of inter-experience relations, making it difficult to resolve conflicts and identify valid co-use combinations. Moreover, as an agent accumulates experience through deployment, memory reliability and relational structure need to be continuously refined to reflect actual performance; yet modifying experience content risks destabilizing previously validated knowledge, motivating a weight-based calibration mechanism that adapts to deployment outcomes without altering what has been learned.

To address these challenges, we propose \textbf{GSEM}, a clinical memory framework that formalizes experiences as \emph{indications} and \emph{contraindications}, with Experience Reliability Validation (ERV) calibrating initial quality $Q$ before deployment, and organizes them into a dual-layer memory graph encoding decision structure within each experience and relational dependencies across experiences, with edge weights $W$ quantifying inter-experience association strength. We further propose an applicability-aware retrieval mechanism combining hybrid seed recall with $Q$ and $W$ guided multi-seed graph traversal, recalling experience combinations that satisfy boundary conditions and exhibit coherent joint-use value. After each task, GSEM self-evolves by updating node quality $Q$ and edge weights $W$ based on outcome feedback, refining memory reliability and relational structure without modifying experience content. Experiments with two backbone LLMs show that GSEM consistently outperforms three categories of strong baselines, achieving 70.90\% and 69.24\% average accuracy with DeepSeek-V3.2 and Qwen3.5-35B respectively, with especially pronounced gains on treatment planning where boundary awareness and multi-step composition matter most.

\vspace{0.3em}
\noindent Our contributions can be summarized as follows:

\noindent(1)~We propose GSEM, a graph-based self-evolving memory framework that formalizes clinical experiences as indications and contraindications, organizing them into a dual-layer memory graph capturing both internal decision structure and inter-experience relations.

\noindent(2)~We design an applicability-aware retrieval mechanism combining hybrid seed recall with LLM-guided multi-seed graph traversal, facilitating boundary-aware selection and relation-aware composition of retrieved experiences.

\noindent(3)~We develop a feedback-driven online calibration method that jointly updates experience quality and relational structure over time, and demonstrate consistent improvements on MedR-Bench and MedAgentsBench under two backbone LLMs.

\section{Related Work}

\subsection{Agent Memory}

Early work treats memory as a managed system resource, paging information between in-context and external memory \citep{packer2023memgpt, kang2025memory}. MemoryBank \citep{zhong2024memorybank} applies an Ebbinghaus forgetting-curve mechanism to selectively reinforce memories, and Mem0 \citep{chhikara2025mem0} dynamically extracts and consolidates key facts to support scalable conversational coherence. ExpeL and AWM \citep{zhao2024expel, wang2024agent} further enable cross-task insight extraction and workflow-level memory without parameter updates. However, most approaches treat memory items as independent entries, leaving relational structure across experiences largely unmodeled.

\subsection{Graph-based Agent Memory}

Several works build graph-based memory representations via knowledge graphs or episodic edges \citep{anokhin2024arigraph, rasmussen2025zep, li2024optimus}. Others focus on graph-enhanced retrieval using dynamic linking or sentence-level structures \citep{xu2025mem, wu2025sgmem}. Mem0$^g$ \citep{chhikara2025mem0} shows graph augmentation yields particular advantages for temporally grounded and multi-hop queries. G-Memory \citep{zhang2025g} introduces hierarchical graph tracing to support simultaneous access to strategic and interaction-level memory. However, these systems offer limited modeling of experience applicability boundaries or inter-experience conflicts.

\subsection{Self-Evolving Memory}

Self-evolving memory enables agents to form and refine memories dynamically from task feedback. Reflexion \citep{shinn2023reflexion} established the foundational paradigm of verbal reflection stored in an episodic buffer to induce better decision-making without weight updates. Building on this, a line of work distills reusable strategies from task trajectories in a closed learning loop \citep{ouyang2025reasoningbank, cao2025remember, cai2025flex}. MemRL \citep{zhang2026memrl} further frames memory evolution as non-parametric reinforcement learning, refining utility estimates over retrieved memories through environmental feedback. MemGen \citep{zhang2025memgen} takes a generative approach, producing latent memories tightly coupled with reasoning. In the medical domain, HealthFlow \citep{zhu2025healthflow} applies self-evolving memory to healthcare agents for meta-level strategic planning. Despite these advances, most existing systems operate on flat memory banks without explicit modeling of relationships across experiences. GSEM bridges this gap by jointly proposing graph-organized experience memory, applicability-aware retrieval, and online structural evolution for clinical reasoning.
\section{Preliminary}
\label{sec:preliminary}

\subsection{Task Definition}

We consider a clinical reasoning agent operating over a sequence of tasks. For each task $T_t$, the agent observes a clinical context $x_t \in \mathcal{X}$, and produces a decision $y_t \in \mathcal{Y}$. The decision function of agent is:
\begin{equation}
    y_t = \pi(x_t,\ \mathcal{R}_t;\, \Theta)
\end{equation}
where $\pi(\cdot;\Theta)$ is parameterized by $\Theta$, and $\mathcal{R}_t$ is a small set of experiences retrieved from the current memory state $\mathcal{M}_t$ that are relevant to $x_t$. After each task, the agent receives a scalar feedback signal and updates $\mathcal{M}_t$ to improve future retrieval.

\subsection{Experience Definition}
\label{sec:experience}

The atomic unit stored in memory is an experience $e_i$, which encodes reusable decision knowledge. Each experience is formally represented as a tuple:
\begin{equation}
    e_i = (c_i,\ s_i,\ z_i,\ Q_i)
\end{equation}
where $c_i \in \mathcal{X}_c$ denotes the applicable \textit{condition} context of the experience, $s_i \in \mathcal{X}_s$ denotes the corresponding decision strategy \textit{content}, $z_i \in \{\oplus, \ominus\}$ is a \textit{polarity label} indicating whether the experience is an \textbf{Indication} ($z_i = \oplus$, a pattern associated with successful outcomes under $c_i$) or a \textbf{Contraindication} ($z_i = \ominus$, a pattern associated with failures under $c_i$ and thus to be avoided), and $Q_i \in [0,1]$ is a scalar quality score measuring the current reliability of the experience.

\subsection{Dual-layer Memory Graph}
\label{sec:memory-graph}

GSEM organizes experiences into a dual-layer memory graph $\mathcal{G}$, which captures both internal decision structure of each experience and the relational knowledge between them.
\begin{equation}
    \mathcal{G} = \bigl(\mathcal{G}^{ent},\ \mathcal{G}^{exp},\ \mathcal{E}^{map}\bigr),
\end{equation}

\paragraph{Entity Layer}
The layer $\mathcal{G}^{ent} = (\mathcal{V}^{ent},\ \mathcal{E}^{ent})$ decomposes each experience into typed decision entities. For each entity node $v \in \mathcal{V}^{ent}$ is assigned a semantic role from $\mathcal{L} = \{\textit{Cond.}, \textit{Constr.}, \textit{Act.}, \textit{Rat.}, \textit{Out.}\}$, capturing the triggering \textit{condition}, applicable \textit{constraints}, chosen \textit{action}, supporting \textit{rationale}, and expected \textit{outcome}, respectively. Directed edges $\mathcal{E}^{ent} \subseteq \mathcal{V}^{ent}\times \mathcal{V}^{ent}$ encode the internal decision flow among entity nodes within each experience.

\paragraph{Experience Layer}
The experience layer $\mathcal{G}^{exp} = (\mathcal{V},\ \mathcal{E},\ Q,\ W)$ encodes relational knowledge between experiences, where $\mathcal{V}=\{e_1,\ldots,e_n\}$ is the set of experience nodes and $\mathcal{E}\subseteq \mathcal{V}\times\mathcal{V}$ is the set of directed inter-experience edges. Each node $e_i$ carries its quality score $Q_i$, and each edge $(e_i,e_j)\in\mathcal{E}$ has a relation weight $W_{ij}\in[0,1]$ indicating the association strength between two experiences.

\paragraph{Cross-layer Mapping}
$\mathcal{E}^{map}\subseteq \mathcal{V}\times\mathcal{V}^{ent}$ links each experience node $e_i$ to its corresponding entity nodes $v$, enabling entity-level structure to inform inter-experience relations.
\section{Methodology}

\begin{figure*}
    \centering
    \includegraphics[width=\textwidth]{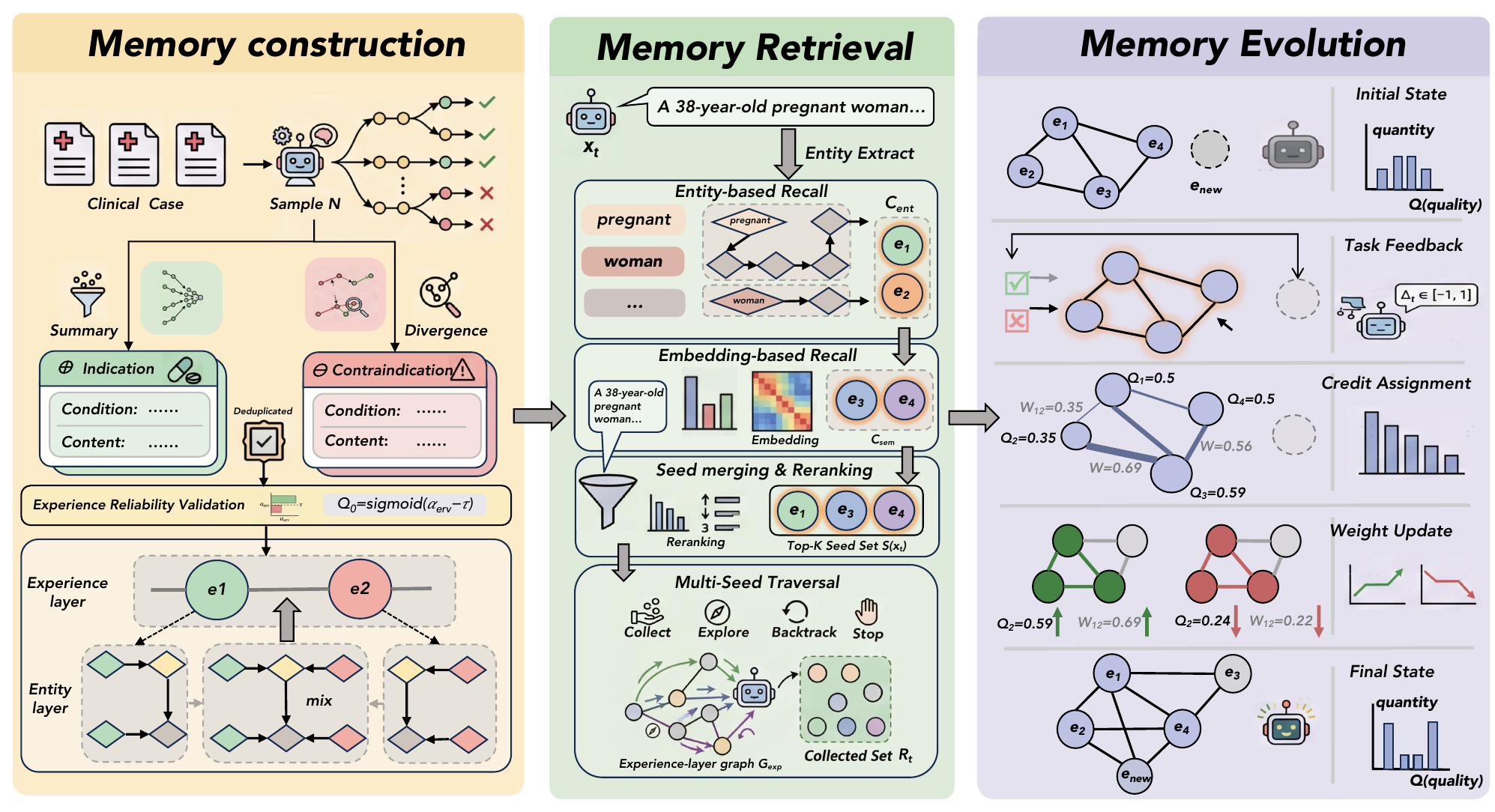}
    \caption{Overall architecture of GSEM, illustrating the three-stage pipeline across Memory Construction, Memory Retrieval, and Memory Evolution.}
    \label{fig:gsem_overview}
\end{figure*}

As shown in Figure~\ref{fig:gsem_overview}, GSEM follows a feedback-driven memory refinement pipeline with three stages. Construction stage (Section~\ref{sec:construction}) extracts structured experiences from historical trajectories and builds the memory graph $\mathcal{G}$. Retrieval stage (Section~\ref{sec:graph-retrieval}) selects task-relevant experiences via LLM-guided graph traversal. Evolution stage (Section~\ref{sec:memory-evolving}) updates the graph to refine future retrieval.

\subsection{Memory Construction}
\label{sec:construction}

\paragraph{Experience Extraction}

For each training instance, we sample $N_{traj}$ reasoning trajectories by stochastic decoding. Successful trajectories are collectively analyzed and summarized into \textit{Indication} experiences, each capturing a generalizable strategy. For each failed trajectory, we choose the most similar successful trajectory and identify a critical divergence through step-wise comparison. The corresponding decision pattern is distilled as a \textit{Contraindication} experience. Finally, we deduplicate near-identical experiences to obtain a compact experience set.

\paragraph{Experience Quality Initialization}

To calibrate the initial reliability of newly extracted experiences, we propose Experience Reliability Validation (ERV). For an experience $e_i$, we inject it into $N_{erv}$ independent held-out trials and record the resulting accuracy $a_{\text{erv}}$. We initialize its quality by
\begin{equation}
    Q_i^{(0)} = \sigma(a_{\text{erv}} - \mu),
\end{equation}
where $\sigma$ denotes the sigmoid function, $\mu = \lceil N_{erv}/2 \rceil / N_{erv}$ represents the baseline performance level, so that experiences performing above $\mu$ will receive higher scores. The score $Q_i^{(0)}$ is used for retrieval and is updated online by task feedback.

\paragraph{Dual-layer Memory Graph Construction}
We organize experiences into the dual-layer memory graph $\mathcal{G}$ (Section~\ref{sec:memory-graph}). Each experience is parsed into core decision entities following a role-based schema to form the entity layer $\mathcal{G}^{ent}$. Then we construct the experience layer $\mathcal{G}^{exp}$ by adding an edge $(e_i,e_j)$ when the similarity of an experience pair
\begin{equation}
    W_{ij}^{(0)} = \alpha S_{\text{entity}} + \beta S_{\text{structure}}
           + \gamma S_{\text{synergy}} + \delta S_{\text{task}}
\end{equation}
exceeds a threshold $\theta_{\text{edge}}$. These four components capture entity overlap, reasoning structure similarity, joint-use value, and task-type consistency, respectively. All components are normalized to $[0,1]$, and the weights are tuned on a validation set.

\subsection{Memory Retrieval}
\label{sec:graph-retrieval}

For an input $x_t$, we retrieve a small set of experiences $\mathcal{R}_t$ via LLM-guided traversal on the experience layer graph $\mathcal{G}^{exp}$. The retriever starts from an initial node set and iteratively alternates between \textit{candidate generation} and \textit{action selection}.

\paragraph{Hybrid Seed Recall}
We generate candidates from two complementary sources. Entity-based recall extracts $\mathrm{Ent}(x_t)$ and retrieves from $\mathcal{G}^{\text{ent}}$ via sparse retrieval, collecting linked experiences as $\mathcal{C}^{ent}$. Embedding-based recall retrieves by dense vector similarity, yielding $\mathcal{C}^{sem}$. The union $\mathcal{C}(x_t) = \mathcal{C}^{ent} \cup \mathcal{C}^{sem}$ is reranked by a linear combination of embedding similarity and BM25 score, and the top-$K$ experiences form the seed set $\mathcal{S}(x_t)$.

\paragraph{Multi-Seed Traversal}
Traversal proceeds simultaneously from all seeds in $\mathcal{S}(x_t)$. At step $h$, for each active path whose current node $e_h$ and let $\mathcal{V}_{vis}$ denote the visited set. We form a forward candidate set by selecting the top-$K$ unvisited neighbors $\mathcal{N}_{K_n}(e_h)=\{e_j\mid(e_h,e_j)\in\mathcal{E}, \,e_j\notin\mathcal{V}_{vis}\}$ ranked by a composite score that combines experience association and node reliability:
\begin{equation}
    \mathrm{score}(e_h,e_j)=\tfrac{1}{2}\bigl(W_{hj}+Q_j\bigr).
\end{equation}
In practice, we set $K_n{=}5$ and only expose the identifier and condition $c_j$ of candidate experience to the agent to keep the decision interface concise.

\paragraph{Action selection}
For each step $h$, the agent chooses one of four discrete actions: \textbf{COLLECT} adds the current experience $e_h$ to $\mathcal{R}_t$; \textbf{EXPLORE} moves to one forward candidate in $\mathcal{N}_{K_n}(e_h)$; \textbf{BACKTRACK} moves to a candidate from an ancestor layer; and \textbf{STOP} terminates retrieval. We enforce a maximum number of traversal steps to prevent unbounded exploration.

\subsection{Self-Evolving Memory}
\label{sec:memory-evolving}

\paragraph{Update Targets}

Memory construction initializes $Q$ and $W$ of $\mathcal{G}^{exp}$ as priors derived from trajectory analysis, but their true utility can only be revealed through real-time task outcomes. Therefore, we evolve the experience-layer memory online using task feedback. Let $\mathcal{G}_t^{exp} =(\mathcal{V}_t,\mathcal{E}_t,Q_t,W_t)$ denote the current state of the experience-layer graph $\mathcal{G}^{exp}_t$, where each node $e_i\in\mathcal{V}_t$ carries a quality $Q_i^{(t)}\in[0,1]$ and each edge $(e_i,e_j)\in\mathcal{E}_t$ carries a relation weight $W_{ij}^{(t)}\in[0,1]$. After completing task $x_t$, it receives a scalar feedback signal $\Delta_t\in[-1,1]$, which will be used to update $\mathcal{G}_t^{exp}$ for improved future retrieval. The state is updated as:
\begin{equation}
    \mathcal{G}_{t+1}^{exp} = \mathcal{F}(\mathcal{G}_t^{exp},\ \Delta_t,\ \mathcal{A}_t)
\end{equation}
where $\mathcal{A}_t = \{e_1, \dots, e_K\}$ is the activated experience set. Only nodes in $\mathcal{A}_t$ and their induced
edges are updated per task.

\paragraph{Evolution Mechanism}
To distribute $\Delta_t$ over activated experiences, we assign each $e_i\in\mathcal{A}_t$ a retrieval rank $r_i$. Node credit is computed by an exponentially decayed normalization:
\begin{equation}
a_i=\frac{\rho^{r_i}}{\sum_{e_j\in\mathcal{A}_t}\rho^{r_j}},\quad \rho\in(0,1)
\end{equation}
where $\rho$ is a decay factor, so that $\sum_{e_i\in\mathcal{A}_t}a_i=1$. Edge credit captures the joint
contribution of co-activated pairs:
\begin{equation}
b_{ij}=\frac{a_i \cdot a_j}{\sum_{(e_k,e_l)\in\mathcal{E}_t^{\mathcal{A}}} a_k \cdot a_l}
\end{equation}

Node quality and relation weights update as:
\begin{equation}
\begin{aligned}
Q_i^{(t+1)} &\leftarrow \mathrm{clip}\bigl(Q_i^{(t)} + \eta_Q\, \cdot a_i\, \cdot \Delta_t,\ 0,\ 1\bigr),\\
\phi_{ij}^{(t+1)} &\leftarrow \phi_{ij}^{(t)} + \eta_W\, \cdot b_{ij}\, \cdot \Delta_t,\\
W_{ij}^{(t+1)} &= \mathrm{clip}\bigl(W_{ij}^{(t)} + \phi_{ij}^{(t+1)},\ 0,\ 1\bigr).
\end{aligned}
\end{equation}
where $\eta_Q$ and $\eta_W$ are the learning rates and $\mathrm{clip}(\cdot)$ enforces the $[0,1]$ range. $W_{ij}^{(t)}$ is prior similarity and $\phi_{ij}$ accumulates the performance history of each edge, ensuring structural priors are refined rather than discarded. This preserves similarity-based structure from construction while gradually calibrating connectivity according to observed outcomes: co-activated pairs that consistently yield positive feedback are strengthened, while those associated with failures are down-weighted.

Beyond updating $(Q,W)$, GSEM can expand coverage by inserting new experiences extracted from recent interactions. Newly added experiences are initialized with $Q^{(0)}$ and connect to existing nodes by reusing the construction-time relation scoring and the same edge threshold, then they are merged into $(\mathcal{V}_t,\mathcal{E}_t)$ for subsequent retrieval.

\section{Experiments}
\label{sec:experiments}

\subsection{Experimental Setup}

\paragraph{Benchmarks}

We evaluate our method on two complementary benchmarks that together cover open-ended clinical reasoning and challenging multi-step inference. \textbf{MedR-Bench}~\citep{MedRBench} contains diagnostic decision-making and treatment planning with open-ended reasoning evaluation. We hold out 329 diagnosis cases and 148 treatment cases for testing. \textbf{MedAgentsBench}~\citep{medagentsbench} is a challenging subset curated from eight medical QA datasets, retaining only questions where fewer than 50\% of models answer correctly. We evaluate on five related subsets: MedBullets, MedQA, MedMCQA, MedExQA, and MedXpertQA. Details are provided in Appendix~\ref{app:benchmarks}.

\begin{table*}[t]
\centering
\small
\begin{tabular}{llcccccccc}
\toprule
\multirow{2}{*}{Model} & \multirow{2}{*}{Method}
& \multicolumn{2}{c}{MedR-Bench}
& \multicolumn{5}{c}{MedAgentsBench} & \multirow{2}{*}{AVG} \\
\cmidrule(lr){3-4}
\cmidrule(lr){5-9}
& & Diag & Treat & MedBullets & MedQA & MedMCQA & MedExQA & MedXpertQA \\
\midrule

\multirow{9}{*}{DeepSeek-V3.2}
& Vanilla        & 87.87 & 57.43 & 48.31 & \underline{74.62} & \underline{40.00} & 28.00 & \underline{31.00} & 64.78 \\
& Na\"{i}ve RAG  & 92.40 & 84.46 & 47.19 & \textbf{74.84} & 38.00 & \underline{29.00} & 27.00 & 68.56 \\
& GraphRAG     & 90.27 & 87.16 & 42.70 & 68.93 & 38.00 & 27.00 & 24.00 & 65.61 \\
\cmidrule(lr){2-10}
& Mem0           & 86.63 & 57.43 & 39.32 & 73.74 & 36.00 & 26.00 & 29.00 & 62.96 \\
& Mem0$^{g}$     & 83.28 & 56.08 & \textbf{50.56} & \underline{74.62} & 37.00 & \underline{29.00} & 27.00 & 63.19 \\
& A-Mem          & \underline{93.62} & \underline{92.57} & 48.31 & 72.43 & 36.00 & 26.00 & \textbf{32.00} & \underline{69.01} \\
\cmidrule(lr){2-10}
& ReMe           & 89.97 & 89.19 & 47.19 & 74.18 & 37.00 & 27.00 & 27.00 & 68.03 \\
& FLEX           & 93.31 & 89.19 & 43.82 & 70.72 & 30.00 & 24.00 & 19.00 & 66.06 \\
\cmidrule(lr){2-10}
& GSEM (ours)    & \textbf{94.22} & \textbf{94.59} & \underline{49.43} & \textbf{74.84} & \textbf{41.00} & \textbf{34.00} & 27.00 & \textbf{70.90} \\

\midrule

\multirow{9}{*}{Qwen3.5-35B}
& Vanilla        & \underline{89.97} & 60.14 & 55.06 & 74.62 & \underline{41.00} & \underline{27.00} & \underline{40.00} & 66.74 \\
& Na\"{i}ve RAG  & 89.06 & 51.35 & 50.56 & 76.37 & \textbf{43.00} & 24.00 & 35.00 & 65.38 \\
& GraphRAG     & 81.16 & 47.30 & 52.81 & 72.00 & \textbf{43.00} & 21.00 & 30.00 & 61.00 \\
\cmidrule(lr){2-10}
& Mem0           & 73.44 & 43.92 & 53.93 & \underline{76.80} & 36.00 & 21.00 & 30.00 & 59.94 \\
& Mem0$^{g}$     & 86.32 & 43.92 & \underline{58.43} & \underline{76.80} & 36.00 &21.00 & \textbf{41.00} & 64.25 \\
& A-Mem          & \underline{89.97} & \underline{65.54} & 52.81 & 71.99 & 38.00 & 23.00 & 36.00 & 65.46 \\
\cmidrule(lr){2-10}
& ReMe           & \textbf{91.49} & 64.86 & 51.69 & 76.38 & 35.00 & 26.00 & 36.00 & \underline{67.20} \\
& FLEX           & 83.59 & 63.51 & 40.45 & 60.61 & 24.00 & 13.00 & 22.00 & 56.01 \\
\cmidrule(lr){2-10}
& GSEM (ours)    & \textbf{91.49} & \textbf{66.89} & \textbf{59.55} & \textbf{77.02} & \underline{41.00} & \textbf{29.00} & \textbf{41.00} & \textbf{69.24}\\

\bottomrule
\end{tabular}
\caption{Main results on MedR-Bench and MedAgentsBench.
\textbf{Acc} denotes outcome accuracy (\%). All MedAgentsBench results report Pass@1 accuracy (\%). Best results are \textbf{bold} and second-best are \underline{underlined}.}
\label{tab:main}
\end{table*}

\paragraph{Baselines}

We compare GSEM with three types of baselines. \textbf{Retrieval-augmented methods} include Naïve RAG~\citep{NaiveRAG} and GraphRAG \citep{Graphrag}. \textbf{Memory-augmented methods} include Mem0~\citep{chhikara2025mem0}, Mem0$^g$, and A-Mem~\citep{xu2025mem}, which augment memory retrieval but do not explicitly model inter-experience relations or applicability boundaries. \textbf{Self-evolving and experience-based methods} include ReMe~\citep{cao2025remember} and FLEX~\citep{cai2025flex}, which represent recent approaches to adaptive clinical memory. All baselines share the same backbone model to ensure fair comparison.

\paragraph{Implementation Details}

All experiments use DeepSeek-V3.2 and Qwen3.5-35B-A3B as the backbone LLMs for generation, retrieval, and judgment, with Qwen-text-embedding-v4 as the embedding model. For memory construction, we sample $N_{traj} = 5$ reasoning trajectories per training instance. For RAG methods, the number of retrieved experiences is set to 5. The edge creation threshold is $\theta_{\text{edge}} = 0.35$. For the self-evolving update, the learning rates for node quality and edge relation weights are $\eta_Q = 0.1$ and $\eta_W = 0.05$ respectively, with rank decay factor $\rho = 0.8$. The initial experience memory is constructed from 50\% diagnosis and 50\% treatment cases sampled from MedR-Bench training split. Full hyperparameter details are provided in Appendix~\ref{app:hyperparameters}.

\paragraph{Metrics}

On MedR-Bench, we report outcome accuracy (\textbf{Acc}) and reasoning quality score (\textbf{R}), where $R$ is the average of three reasoning evaluation dimensions: efficiency, factuality, and completeness. Following the benchmark protocol, we use DeepSeek-V3.2 as the LLM judge to score these reasoning dimensions based on the generated rationale. On MedAgentsBench, we report Pass@1 accuracy across all five subsets.

\subsection{Main Results}

Table~\ref{tab:main} summarizes the main results on MedR-Bench and MedAgentsBench under two backbone models. On MedR-Bench, GSEM achieves the best overall accuracy with DeepSeek-V3.2, reaching 94.22\% on diagnosis and 94.59\% on treatment. The improvements are consistent across baseline families: compared with RAG-style methods, GSEM substantially improves treatment accuracy (94.59\% vs. 87.16\%), suggesting that modeling inter-experience relations and enabling structured traversal is beneficial beyond retrieving independent snippets. Compared with memory-augmented baselines, GSEM further improves upon the strongest competitor A-Mem (94.59\% vs. 92.57\%), indicates that similarity-based retrieval over a flat memory is insufficient---explicitly modeling applicability boundaries and inter-experience relations in a structured graph enables more precise and coherent experience selection. GSEM also outperforms self-evolving baselines on both tasks, demonstrating robust gains in experience-intensive reasoning settings.

On MedAgentsBench, GSEM performs competitively overall and achieves
the highest average accuracy across subsets under both backbones
(70.90\% with DeepSeek-V3.2 and 69.24\% with Qwen3.5-35B). Performance varies across
individual subsets, which we attribute primarily to data contamination
in recent large-scale pretrained models: on subsets where the backbone
has likely encountered similar questions during pretraining, parametric
recall dominates and diminishes the marginal benefit of externally
retrieved experiences. This effect is most pronounced on expert-level
subsets such as MedXpertQA, where even the vanilla baseline achieves
competitive scores, suggesting that experience-augmented methods
provide the strongest gains when the task genuinely requires
experiential reasoning beyond parametric knowledge.

\subsection{Self-evolving Analysis}

\paragraph{Evolution Effectiveness}
To evaluate the self-evolution (SE) mechanism, we test GSEM at three snapshots after $50$, $150$, and $250$ online evolution updates, and compare them with the static variant without evolution. As shown in Table~\ref{tab:self_evolve}, all evolved snapshots outperform the base model on both diagnosis and treatment tasks. After only 50 updates, self-evolving reduces the number of incorrect cases from 19 to 9 for diagnosis and from 8 to 4 for treatment, yielding a substantial error reduction at an already strong baseline. While the magnitude of improvement varies across snapshots, the gains persist throughout the evolution process, suggesting that self-evolution provides consistent benefits rather than one-off fluctuations.

\paragraph{Memory Calibration Dynamics}
Beyond accuracy gains, we examine how self-evolving mechanism reshapes the experience-layer memory through changes in node quality $Q$ and relation weight $W$ (Figure~\ref{fig:Q_cropped} and Figure~\ref{fig:W_cropped}). After evolution, the distribution of $Q$ becomes more polarized: experiences that repeatedly support correct decisions receive positive feedback and move toward higher quality, whereas underperforming experiences are down-weighted. This increased separation helps retrieval prioritize high-utility experiences. Meanwhile, the distribution of $W$ becomes more selective: relations between co-activated experiences that correlate with successful outcomes are strengthened, while relations frequently involved in failures are weakened. Overall, self-evolving progressively recalibrates both memory content ($Q$) and relational structure ($W$), shifting the graph from a similarity-driven prior toward a performance-informed topology that better supports future retrieval.

\begin{table}[t]
\centering
\small
\begin{tabular}{lccccccc}
\toprule
\multirow{2}{*}{Method}
& \multicolumn{2}{c}{MedR-Bench} \\
\cmidrule(lr){2-3}
& Diagnosis / Acc & Treatment / Acc \\
\midrule
GSEM        & 94.22 & 94.59 \\
w/ SE(50)   & 97.26 & \textbf{97.30} \\
w/ SE(150)  & \textbf{97.87} & 95.95 \\
w/ SE(250)  & 96.96 & \textbf{97.30} \\
\bottomrule
\end{tabular}
\caption{Effect of the number of online evolution updates on GSEM performance. Results are reported after online memory evolution using different numbers of samples. \textbf{Bold} indicates the best performance.}
\label{tab:self_evolve}
\end{table}
\section{Further Analysis}

\subsection{Retrieval Ablation Study}

To quantify the contribution of each component in our hybrid retriever, we ablate entity-based recall, embedding-based recall, and multi-seed traversal. Table~\ref{tab:ablation} reports results on a representative subset of benchmarks. Specifically, we include MedR-Bench to represent open-ended clinical reasoning, and MedBullets to evaluate closed-form inference under complex and condition-intensive settings. Removing entity-based recall mainly hurts the condition-intensive setting, confirming its role in anchoring retrieval to decision-relevant conditions. Removing embedding-based recall causes the largest drops, especially on Treatment and MedBullets, highlighting the need for dense semantic matching under diverse surface forms. Replacing multi-seed with single-seed traversal consistently degrades performance, suggesting that multiple entry points improve graph coverage and reduce sensitivity to initialization.

\begin{table}[t]
\centering
\small
\setlength{\tabcolsep}{6pt}
\begin{tabular}{lcccc}
\toprule
\textbf{Setting} & \textbf{Diag.} & \textbf{Treat.} & \textbf{MedBullets}\\
\midrule
w/o entity\_recall         & 92.71 & 92.57  & 24.00\\
w/o embedding\_recall      & 93.62 & 83.78  & 10.00\\
w/o multi\_seed\_retrieval  & 91.79 & 92.57  & 23.00\\
\midrule
GSEM                       & \textbf{94.22} & \textbf{94.59} & \textbf{34.00}\\
\bottomrule
\end{tabular}
\caption{Ablation results on diagnosis (Diag.) / treatment (Treat.) accuracy and QA benchmark MedBullets. \textbf{Bold} indicates the best performance per task.}
\label{tab:ablation}
\end{table}

\begin{figure*}[t]
    \centering
    \includegraphics[width=\textwidth]{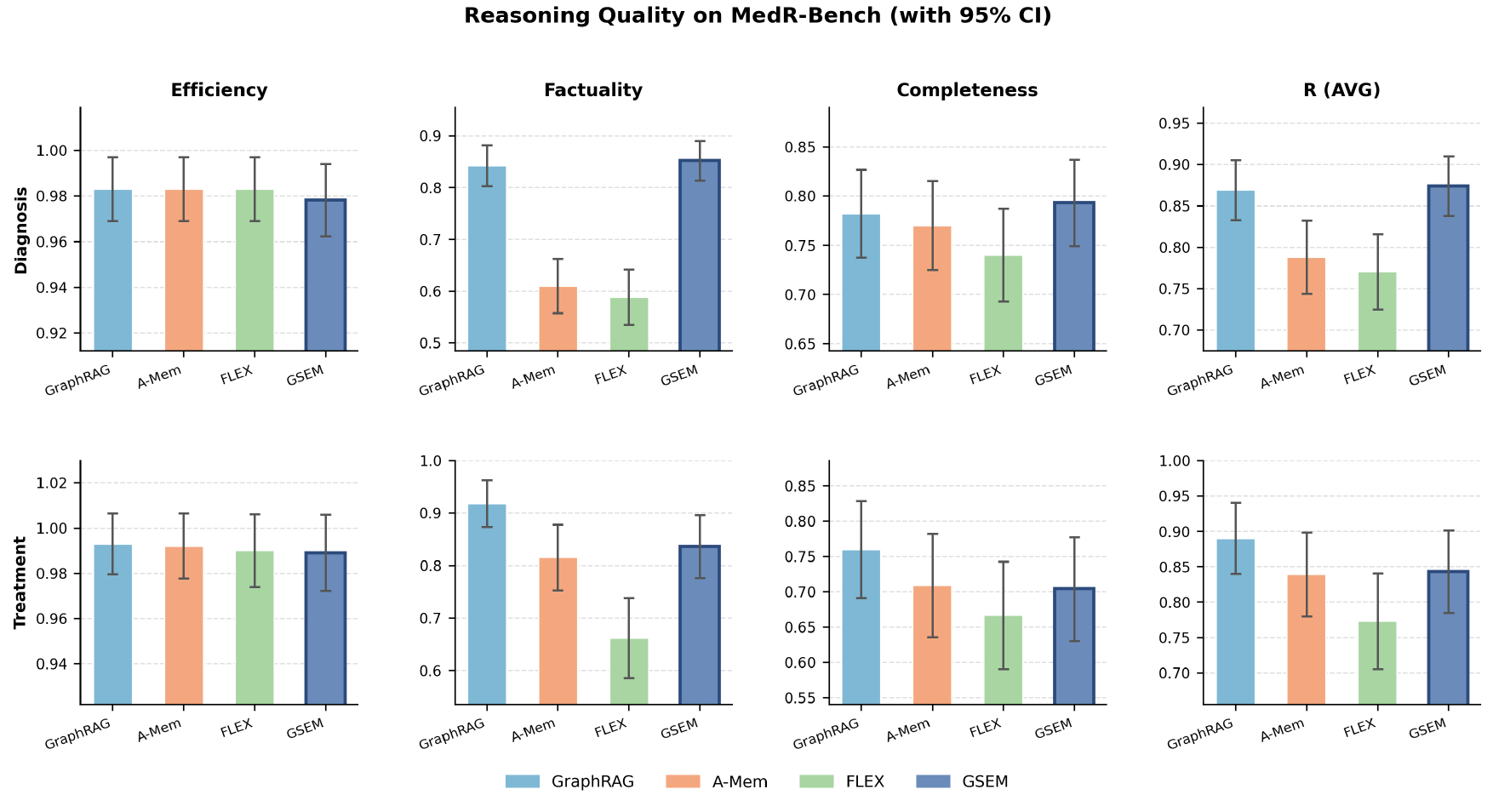}
    \caption{Per-dimension reasoning quality (Efficiency, Factuality,
    Completeness, and R (AVG)) on MedR-Bench diagnosis and treatment
    tasks. Error bars denote two-sided 95\% confidence intervals.}
    \label{fig:reasoning_quality}
\end{figure*}

\subsection{Reasoning Ability Analysis}

We compare GSEM against strong representatives from each baseline category on MedR-Bench: GraphRAG (RAG-style), A-Mem (memory-augmented), and FLEX (self-evolving). Table~\ref{tab:reasoning} reports the aggregated reasoning quality score $R$ ($mean_{\pm CI}$) for diagnosis and treatment, and Figure~\ref{fig:reasoning_quality} further breaks down $R$ into Efficiency, Factuality, and Completeness (details with 95\% confidence intervals are provided in Appendix~\ref{app:reasoning_detail}).

Notably, GSEM achieves the best Diagnosis/R, and while GraphRAG attains the highest Treatment/R, its treatment accuracy is substantially lower, suggesting that reasoning-quality scores do not necessarily translate into more reliable final decisions. Overall, GSEM is the only method that consistently performs well on both reasoning quality and task accuracy, indicating that structured experience retrieval improves not only decision correctness but also the factual grounding and completeness of the reasoning process.

\begin{table}[t]
\centering
\small
\begin{tabular}{lccccccc}
\toprule
\multirow{2}{*}{Method}
& \multicolumn{2}{c}{MedR-Bench} \\
\cmidrule(lr){2-3}
& Diagnosis / R & Treatment / R \\
\midrule
GraphRAG  & $86.9_{\pm3.6}$ & \textbf{$\mathbf{89.0}_{\pm5.0}$} \\
A-Mem     & $78.8_{\pm4.4}$ & $83.9_{\pm5.9}$ \\
FLEX      & $77.0_{\pm4.5}$ & $77.3_{\pm6.7}$  \\
GSEM      & $\mathbf{87.4}_{\pm3.6}$ & $84.3_{\pm5.9}$  \\

\bottomrule
\end{tabular}
\caption{Reasoning quality on MedR-Bench (mean$\pm$CI). R denotes the aggregated reasoning quality score. \textbf{Bold} indicates the best performance.}
\label{tab:reasoning}
\end{table}

\subsection{Effect of Model Size Analysis}

\begin{table}[t]
\centering
\small
\begin{tabular}{lccccccc}
\toprule
Retriever + Generator & Diag. / Acc & Treat. / Acc \\
\midrule
Qwen + Qwen        &91.49  &66.89  \\
Qwen + DeepSeek   &\textbf{96.66}  &\textbf{95.95}  \\
DeepSeek + Qwen  &87.84  &61.49  \\
DeepSeek + DeepSeek  &94.22  &94.59  \\

\bottomrule
\end{tabular}
\caption{Effect of model size on GSEM by swapping the generation model and the retrieval model. \textbf{Bold} indicates the best performance.}
\label{tab:model_size}
\end{table}

We study the effect of model size in GSEM by swapping the retriever and the generator between DeepSeek-V3.2 (671B) and Qwen3.5 (35B), resulting in four configurations (Table~\ref{tab:model_size}). Overall, the results exhibit a clear asymmetry: performance is largely determined by the generator, while the retriever choice has a comparatively smaller effect. When the generator is replaced with a smaller model, accuracy drops substantially, particularly on Treatment/Acc. This observation is consistent with treatment planning requiring sustained multi-step reasoning that is more sensitive to generation capacity. In contrast, diagnosis is relatively more robust, likely because it relies more on matching structured clinical findings than on long-horizon strategy synthesis. Notably, pairing a strong generator with a lightweight retriever does not degrade performance in our setting (Qwen+DeepSeek performs best), and the difference to using a large retriever is small. These results suggest that our graph-based retrieval can be executed effectively with a smaller retriever, enabling more cost-efficient deployment.

\section{Conclusion}
\label{sec:conclusion}

We studied how to equip clinical LLM agents with structured experience memory for reliable reuse of past decision knowledge. To this end, we introduced GSEM, which formalizes clinical experiences as indications and contraindications, organizes them into a dual-layer memory graph encoding decision structure and inter-experience relational dependencies, and supports applicability-aware retrieval via condition-compatible seed recall and graph traversal. After each task, GSEM self-evolves by calibrating memory reliability and relational structure based on outcome feedback, without modifying experience content. Experiments show that GSEM consistently outperforms three categories of baselines, with particularly strong gains on multi-step treatment planning. Future work includes incorporating richer feedback signals, strengthening safety and attribution under clinical constraints, and evaluating memory evolution in longer-horizon real world workflows.

\section*{Limitations}

This study has several limitations that could be addressed in future work. First, we evaluate GSEM on public benchmarks with automatic protocols. While these benchmarks cover open-ended reasoning and challenging QA, they do not fully reflect real clinical workflows, interactive decision processes, longitudinal follow-up, or deployment constraints. Additional validation in more realistic settings, ideally with expert assessment, is therefore necessary. Secondly, memory construction requires sampling multiple trajectories and performing reliability validation, which increases computation. As the memory graph grows, maintaining links and performing graph traversal may add latency. Finally, our retriever relies on an LLM-guided traversal policy with discrete actions. Although this design enables controllable, compositional retrieval, it can be sensitive to prompting and introduce non-determinism in path selection. Future work could explore more stable retrieval policies and stronger constraints for safety-critical use, for example by learning a retrieval policy from feedback.
\section*{Ethical Consideration}

This work studies continual memory in clinical reasoning agents and is not intended for direct clinical use. In high-stakes settings, erroneous recommendations may cause harm if its outputs are over-trusted or used without qualified human oversight. To mitigate these risks, any real-world use should incorporate human-in-the-loop verification. Our online memory evolution uses task-level accuracy as the feedback signal. While simple and objective, accuracy is a coarse-grained metric and may not capture safety-critical nuances, so deployment should impose additional safeguards and monitoring to prevent unsafe drift. We evaluate on public benchmarks and do not collect identifiable patient information in our experiments.

\bibliography{latex/main}

@article{packer2023memgpt,
  title={MemGPT: towards LLMs as operating systems.},
  author={Packer, Charles and Fang, Vivian and Patil, Shishir\_G and Lin, Kevin and Wooders, Sarah and Gonzalez, Joseph\_E},
  year={2023},
  publisher={ArXiv}
}

@inproceedings{kang2025memory,
  title={Memory os of ai agent},
  author={Kang, Jiazheng and Ji, Mingming and Zhao, Zhe and Bai, Ting},
  booktitle={Proceedings of the 2025 Conference on Empirical Methods in Natural Language Processing},
  pages={25972--25981},
  year={2025}
}

@inproceedings{zhong2024memorybank,
  title={Memorybank: Enhancing large language models with long-term memory},
  author={Zhong, Wanjun and Guo, Lianghong and Gao, Qiqi and Ye, He and Wang, Yanlin},
  booktitle={Proceedings of the AAAI conference on artificial intelligence},
  volume={38},
  number={17},
  pages={19724--19731},
  year={2024}
}

@article{chhikara2025mem0,
  title={Mem0: Building production-ready ai agents with scalable long-term memory},
  author={Chhikara, Prateek and Khant, Dev and Aryan, Saket and Singh, Taranjeet and Yadav, Deshraj},
  journal={arXiv preprint arXiv:2504.19413},
  year={2025}
}

@inproceedings{zhao2024expel,
  title={Expel: Llm agents are experiential learners},
  author={Zhao, Andrew and Huang, Daniel and Xu, Quentin and Lin, Matthieu and Liu, Yong-Jin and Huang, Gao},
  booktitle={Proceedings of the AAAI Conference on Artificial Intelligence},
  volume={38},
  number={17},
  pages={19632--19642},
  year={2024}
}

@article{wang2024agent,
  title={Agent workflow memory},
  author={Wang, Zora Zhiruo and Mao, Jiayuan and Fried, Daniel and Neubig, Graham},
  journal={arXiv preprint arXiv:2409.07429},
  year={2024}
}

@article{anokhin2024arigraph,
  title={Arigraph: Learning knowledge graph world models with episodic memory for llm agents},
  author={Anokhin, Petr and Semenov, Nikita and Sorokin, Artyom and Evseev, Dmitry and Kravchenko, Andrey and Burtsev, Mikhail and Burnaev, Evgeny},
  journal={arXiv preprint arXiv:2407.04363},
  year={2024}
}

@article{rasmussen2025zep,
  title={Zep: a temporal knowledge graph architecture for agent memory},
  author={Rasmussen, Preston and Paliychuk, Pavlo and Beauvais, Travis and Ryan, Jack and Chalef, Daniel},
  journal={arXiv preprint arXiv:2501.13956},
  year={2025}
}

@article{li2024optimus,
  title={Optimus-1: Hybrid multimodal memory empowered agents excel in long-horizon tasks},
  author={Li, Zaijing and Xie, Yuquan and Shao, Rui and Chen, Gongwei and Jiang, Dongmei and Nie, Liqiang},
  journal={Advances in neural information processing systems},
  volume={37},
  pages={49881--49913},
  year={2024}
}

@article{xu2025mem,
  title={A-mem: Agentic memory for llm agents},
  author={Xu, Wujiang and Liang, Zujie and Mei, Kai and Gao, Hang and Tan, Juntao and Zhang, Yongfeng},
  journal={arXiv preprint arXiv:2502.12110},
  year={2025}
}

@article{wu2025sgmem,
  title={Sgmem: Sentence graph memory for long-term conversational agents},
  author={Wu, Yaxiong and Zhang, Yongyue and Liang, Sheng and Liu, Yong},
  journal={arXiv preprint arXiv:2509.21212},
  year={2025}
}

@article{zhang2025g,
  title={G-memory: Tracing hierarchical memory for multi-agent systems},
  author={Zhang, Guibin and Fu, Muxin and Wan, Guancheng and Yu, Miao and Wang, Kun and Yan, Shuicheng},
  journal={arXiv preprint arXiv:2506.07398},
  year={2025}
}

@article{shinn2023reflexion,
  title={Reflexion: Language agents with verbal reinforcement learning},
  author={Shinn, Noah and Cassano, Federico and Gopinath, Ashwin and Narasimhan, Karthik and Yao, Shunyu},
  journal={Advances in neural information processing systems},
  volume={36},
  pages={8634--8652},
  year={2023}
}

@article{ouyang2025reasoningbank,
  title={Reasoningbank: Scaling agent self-evolving with reasoning memory},
  author={Ouyang, Siru and Yan, Jun and Hsu, I and Chen, Yanfei and Jiang, Ke and Wang, Zifeng and Han, Rujun and Le, Long T and Daruki, Samira and Tang, Xiangru and others},
  journal={arXiv preprint arXiv:2509.25140},
  year={2025}
}

@article{cao2025remember,
  title={Remember me, refine me: A dynamic procedural memory framework for experience-driven agent evolution},
  author={Cao, Zouying and Deng, Jiaji and Yu, Li and Zhou, Weikang and Liu, Zhaoyang and Ding, Bolin and Zhao, Hai},
  journal={arXiv preprint arXiv:2512.10696},
  year={2025}
}

@article{cai2025flex,
  title={Flex: Continuous agent evolution via forward learning from experience},
  author={Cai, Zhicheng and Guo, Xinyuan and Pei, Yu and Feng, Jiangtao and Su, Jinsong and Chen, Jiangjie and Zhang, Ya-Qin and Ma, Wei-Ying and Wang, Mingxuan and Zhou, Hao},
  journal={arXiv preprint arXiv:2511.06449},
  year={2025}
}

@article{zhang2026memrl,
  title={Memrl: Self-evolving agents via runtime reinforcement learning on episodic memory},
  author={Zhang, Shengtao and Wang, Jiaqian and Zhou, Ruiwen and Liao, Junwei and Feng, Yuchen and Li, Zhuo and Zheng, Yujie and Zhang, Weinan and Wen, Ying and Li, Zhiyu and others},
  journal={arXiv preprint arXiv:2601.03192},
  year={2026}
}

@article{zhang2025memgen,
  title={Memgen: Weaving generative latent memory for self-evolving agents},
  author={Zhang, Guibin and Fu, Muxin and Yan, Shuicheng},
  journal={arXiv preprint arXiv:2509.24704},
  year={2025}
}

@article{zhu2025healthflow,
  title={HealthFlow: A Self-Evolving AI Agent with Meta Planning for Autonomous Healthcare Research},
  author={Zhu, Yinghao and Qi, Yifan and Wang, Zixiang and Gu, Lei and Sui, Dehao and Hu, Haoran and Zhang, Xichen and He, Ziyi and He, Junjun and Ma, Liantao and others},
  journal={arXiv preprint arXiv:2508.02621},
  year={2025}
}

@article{MedRBench,
  title={Quantifying the reasoning abilities of llms on real-world clinical cases},
  author={Qiu, Pengcheng and Wu, Chaoyi and Liu, Shuyu and Zhao, Weike and Chen, Zhuoxia and Gu, Hongfei and Peng, Chuanjin and Zhang, Ya and Wang, Yanfeng and Xie, Weidi},
  journal={arXiv preprint arXiv:2503.04691},
  year={2025}
}

@article{medagentsbench,
  title={Medagentsbench: Benchmarking thinking models and agent frameworks for complex medical reasoning},
  author={Tang, Xiangru and Shao, Daniel and Sohn, Jiwoong and Chen, Jiapeng and Zhang, Jiayi and Xiang, Jinyu and Wu, Fang and Zhao, Yilun and Wu, Chenglin and Shi, Wenqi and others},
  journal={arXiv preprint arXiv:2503.07459},
  year={2025}
}

@Article{MedQA,
AUTHOR = {Jin, Di and Pan, Eileen and Oufattole, Nassim and Weng, Wei-Hung and Fang, Hanyi and Szolovits, Peter},
TITLE = {What Disease Does This Patient Have? A Large-Scale Open Domain Question Answering Dataset from Medical Exams},
JOURNAL = {Applied Sciences},
VOLUME = {11},
YEAR = {2021},
NUMBER = {14},
ARTICLE-NUMBER = {6421},
URL = {https://www.mdpi.com/2076-3417/11/14/6421},
ISSN = {2076-3417},
DOI = {10.3390/app11146421}
}

@InProceedings{MedMCQA,
  title = 	 {MedMCQA: A Large-scale Multi-Subject Multi-Choice Dataset for Medical domain Question Answering},
  author =       {Pal, Ankit and Umapathi, Logesh Kumar and Sankarasubbu, Malaikannan},
  booktitle = 	 {Proceedings of the Conference on Health, Inference, and Learning},
  pages = 	 {248--260},
  year = 	 {2022},
  editor = 	 {Flores, Gerardo and Chen, George H and Pollard, Tom and Ho, Joyce C and Naumann, Tristan},
  volume = 	 {174},
  series = 	 {Proceedings of Machine Learning Research},
  month = 	 {07--08 Apr},
  publisher =    {PMLR},
  url = 	 {https://proceedings.mlr.press/v174/pal22a.html}
}

@inproceedings{Medbullets,
    title = "Benchmarking Large Language Models on Answering and Explaining Challenging Medical Questions",
    author = "Chen, Hanjie  and
      Fang, Zhouxiang  and
      Singla, Yash  and
      Dredze, Mark",
    editor = "Chiruzzo, Luis  and
      Ritter, Alan  and
      Wang, Lu",
    booktitle = "Proceedings of the 2025 Conference of the Nations of the Americas Chapter of the Association for Computational Linguistics: Human Language Technologies (Volume 1: Long Papers)",
    month = apr,
    year = "2025",
    address = "Albuquerque, New Mexico",
    publisher = "Association for Computational Linguistics",
    url = "https://aclanthology.org/2025.naacl-long.182/",
    doi = "10.18653/v1/2025.naacl-long.182",
    pages = "3563--3599",
    ISBN = "979-8-89176-189-6"
}

@InProceedings{MedXpertQA,
  title = 	 {{M}ed{X}pert{QA}: Benchmarking Expert-Level Medical Reasoning and Understanding},
  author =       {Zuo, Yuxin and Qu, Shang and Li, Yifei and Chen, Zhang-Ren and Zhu, Xuekai and Hua, Ermo and Zhang, Kaiyan and Ding, Ning and Zhou, Bowen},
  booktitle = 	 {Proceedings of the 42nd International Conference on Machine Learning},
  pages = 	 {80961--80990},
  year = 	 {2025},
  editor = 	 {Singh, Aarti and Fazel, Maryam and Hsu, Daniel and Lacoste-Julien, Simon and Berkenkamp, Felix and Maharaj, Tegan and Wagstaff, Kiri and Zhu, Jerry},
  volume = 	 {267},
  series = 	 {Proceedings of Machine Learning Research},
  month = 	 {13--19 Jul},
  publisher =    {PMLR},
  url = 	 {https://proceedings.mlr.press/v267/zuo25a.html}
}

@misc{MedExQA,
      title={MedExQA: Medical Question Answering Benchmark with Multiple Explanations}, 
      author={Yunsoo Kim and Jinge Wu and Yusuf Abdulle and Honghan Wu},
      year={2024},
      eprint={2406.06331},
      archivePrefix={arXiv},
      primaryClass={cs.CL},
      url={https://arxiv.org/abs/2406.06331}, 
}

@inproceedings{NaiveRAG,
    title = "Dense Passage Retrieval for Open-Domain Question Answering",
    author = "Karpukhin, Vladimir  and
      Oguz, Barlas  and
      Min, Sewon  and
      Lewis, Patrick  and
      Wu, Ledell  and
      Edunov, Sergey  and
      Chen, Danqi  and
      Yih, Wen-tau",
    editor = "Webber, Bonnie  and
      Cohn, Trevor  and
      He, Yulan  and
      Liu, Yang",
    booktitle = "Proceedings of the 2020 Conference on Empirical Methods in Natural Language Processing (EMNLP)",
    month = nov,
    year = "2020",
    address = "Online",
    publisher = "Association for Computational Linguistics",
    url = "https://aclanthology.org/2020.emnlp-main.550/",
    doi = "10.18653/v1/2020.emnlp-main.550",
    pages = "6769--6781"
}

@misc{Graphrag,
      title={From Local to Global: A Graph RAG Approach to Query-Focused Summarization}, 
      author={Darren Edge and Ha Trinh and Newman Cheng and Joshua Bradley and Alex Chao and Apurva Mody and Steven Truitt and Dasha Metropolitansky and Robert Osazuwa Ness and Jonathan Larson},
      year={2025},
      eprint={2404.16130},
      archivePrefix={arXiv},
      primaryClass={cs.CL},
      url={https://arxiv.org/abs/2404.16130}, 
}

@inproceedings{medagents,
    title = "{M}ed{A}gents: Large Language Models as Collaborators for Zero-shot Medical Reasoning",
    author = "Tang, Xiangru  and
      Zou, Anni  and
      Zhang, Zhuosheng  and
      Li, Ziming  and
      Zhao, Yilun  and
      Zhang, Xingyao  and
      Cohan, Arman  and
      Gerstein, Mark",
    editor = "Ku, Lun-Wei  and
      Martins, Andre  and
      Srikumar, Vivek",
    booktitle = "Findings of the Association for Computational Linguistics: ACL 2024",
    month = aug,
    year = "2024",
    address = "Bangkok, Thailand",
    publisher = "Association for Computational Linguistics",
    url = "https://aclanthology.org/2024.findings-acl.33/",
    doi = "10.18653/v1/2024.findings-acl.33",
    pages = "599--621"
}

@inproceedings{MDAgents,
 author = {Kim, Yubin and Park, Chanwoo and Jeong, Hyewon and Chan, Yik Siu and Xu, Xuhai and McDuff, Daniel and Lee, Hyeonhoon and Ghassemi, Marzyeh and Breazeal, Cynthia and Park, Hae Won},
 booktitle = {Advances in Neural Information Processing Systems},
 doi = {10.52202/079017-2522},
 editor = {A. Globerson and L. Mackey and D. Belgrave and A. Fan and U. Paquet and J. Tomczak and C. Zhang},
 pages = {79410--79452},
 publisher = {Curran Associates, Inc.},
 title = {MDAgents: An Adaptive Collaboration of LLMs for Medical Decision-Making},
 url = {https://proceedings.neurips.cc/paper_files/paper/2024/file/90d1fc07f46e31387978b88e7e057a31-Paper-Conference.pdf},
 volume = {37},
 year = {2024}
}

@inproceedings{MediQ,
 author = {Li, Shuyue Stella and Balachandran, Vidhisha and Feng, Shangbin and Ilgen, Jonathan S. and Pierson, Emma and Koh, Pang Wei and Tsvetkov, Yulia},
 booktitle = {Advances in Neural Information Processing Systems},
 doi = {10.52202/079017-0908},
 editor = {A. Globerson and L. Mackey and D. Belgrave and A. Fan and U. Paquet and J. Tomczak and C. Zhang},
 pages = {28858--28888},
 publisher = {Curran Associates, Inc.},
 title = {MediQ: Question-Asking LLMs and a Benchmark for Reliable Interactive Clinical Reasoning},
 url = {https://proceedings.neurips.cc/paper_files/paper/2024/file/32b80425554e081204e5988ab1c97e9a-Paper-Conference.pdf},
 volume = {37},
 year = {2024}
}

@inproceedings{selfrag,
title={Self-{RAG}: Learning to Retrieve, Generate, and Critique through Self-Reflection},
author={Akari Asai and Zeqiu Wu and Yizhong Wang and Avirup Sil and Hannaneh Hajishirzi},
booktitle={The Twelfth International Conference on Learning Representations},
year={2024},
url={https://openreview.net/forum?id=hSyW5go0v8}
}

@inproceedings{rarr,
    title = "{RARR}: Researching and Revising What Language Models Say, Using Language Models",
    author = "Gao, Luyu  and
      Dai, Zhuyun  and
      Pasupat, Panupong  and
      Chen, Anthony  and
      Chaganty, Arun Tejasvi  and
      Fan, Yicheng  and
      Zhao, Vincent  and
      Lao, Ni  and
      Lee, Hongrae  and
      Juan, Da-Cheng  and
      Guu, Kelvin",
    editor = "Rogers, Anna  and
      Boyd-Graber, Jordan  and
      Okazaki, Naoaki",
    booktitle = "Proceedings of the 61st Annual Meeting of the Association for Computational Linguistics (Volume 1: Long Papers)",
    month = jul,
    year = "2023",
    address = "Toronto, Canada",
    publisher = "Association for Computational Linguistics",
    url = "https://aclanthology.org/2023.acl-long.910/",
    doi = "10.18653/v1/2023.acl-long.910",
    pages = "16477--16508"
}

\appendix

\section{Benchmark Details}
\label{app:benchmarks}

\paragraph{MedR-Bench}
MedR-Bench \citep{MedRBench} is a benchmark designed to evaluate the medical reasoning capabilities of large language models using real-world clinical cases. It focuses on multi-step clinical decision making, including examination recommendation, diagnostic reasoning, and treatment planning, rather than isolated question answering. The benchmark consists of structured patient cases derived from clinical reports and supports both final-outcome evaluation and process-level reasoning assessment. In our experiments, a model is evaluated based on its ability to produce clinically plausible decisions along with factually grounded and efficient reasoning under different information availability settings.

\paragraph{MedAgentsBench}
MedAgentsBench \citep{medagentsbench} is a benchmark that evaluates large language models and agent-based systems on challenging medical question answering tasks that require complex, multi-step clinical reasoning. It aggregates questions from multiple established medical datasets and applies difficulty filtering to emphasize cases where simple prompting methods are insufficient. Evaluation is conducted in a zero-shot setting using Pass@1 accuracy, enabling consistent comparison across base models, reasoning strategies, and agent workflows. In this work, we use MedAgentsBench to assess the effectiveness of our method on complex exam-style medical reasoning tasks.

\begin{itemize}[nosep,leftmargin=*]
    \item \textbf{MedQA} \citep{MedQA} A medical multiple-choice question answering benchmark constructed from professional medical licensing examinations. It primarily evaluates general medical knowledge and exam-style clinical reasoning.
    \item \textbf{MedMCQA} \citep{MedMCQA} A large-scale, multi-subject medical QA dataset derived from postgraduate medical entrance examinations. It covers a broad range of medical specialties and assesses both factual knowledge and conceptual understanding.
    \item \textbf{Medbullets} \citep{Medbullets} A collection of clinically oriented multiple-choice questions modeled after USMLE Step 2/3 exams. The questions emphasize realistic clinical scenarios and are accompanied by expert-written explanations.
    \item \textbf{MedExQA} \citep{MedExQA} A medical QA benchmark designed to evaluate models’ understanding of medical knowledge through explanation generation. Each question is paired with multiple reference explanations, enabling assessment beyond answer accuracy.
    \item \textbf{MedXpertQA} \citep{MedXpertQA} A challenging benchmark targeting expert-level medical knowledge and advanced reasoning. It includes both text-only and multimodal questions sourced from high-difficulty medical and specialty board examinations.
\end{itemize}

\section{Baselines Details}

\paragraph{Na\"{i}ve RAG}
Retrieval-Augmented Generation (RAG) is implemented using a dual-retrieval pipeline that combines sparse and dense retrieval \citep{NaiveRAG}. Specifically, we adopt a BM25-based lexical retriever together with a dense vector retriever based on contextual embeddings, following the design principles of Dense Passage Retrieval. For each query, candidate documents retrieved by BM25 and embedding-based similarity search are merged to form the final retrieval set. The retrieved experiences are concatenated and provided as context to LLM for answer generation. This baseline serves as a strong and widely adopted RAG approach that balances exact keyword matching and semantic retrieval.

\paragraph{GraphRAG}
GraphRAG \citep{Graphrag} extends standard RAG by organizing retrieved knowledge into an explicit graph structure that captures entities and their relations. It constructs a knowledge graph from the corpus and performs graph-informed retrieval to identify relevant subgraphs conditioned on the query. Instead of treating retrieved documents as independent passages, GraphRAG leverages relational context and global structural information to support more coherent reasoning. The retrieved graph-based context is then linearized and provided to LLM for generation. The baseline represents structure-aware retrieval and is designed to alleviate the limitations of flat document retrieval in complex reasoning scenarios.

\begin{figure*}[t]
    \centering
    \includegraphics[width=\linewidth]{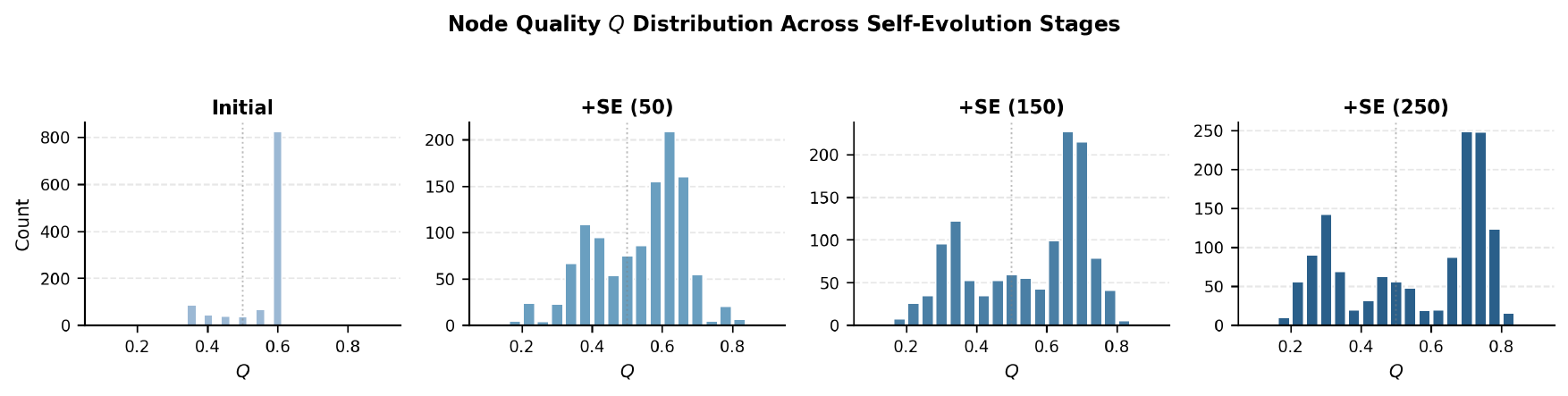}
    \caption{Distribution of node quality $Q$ in self-evolving stage.}
    \label{fig:Q_cropped}
\end{figure*}

\begin{figure*}[t]
    \centering
    \includegraphics[width=\linewidth]{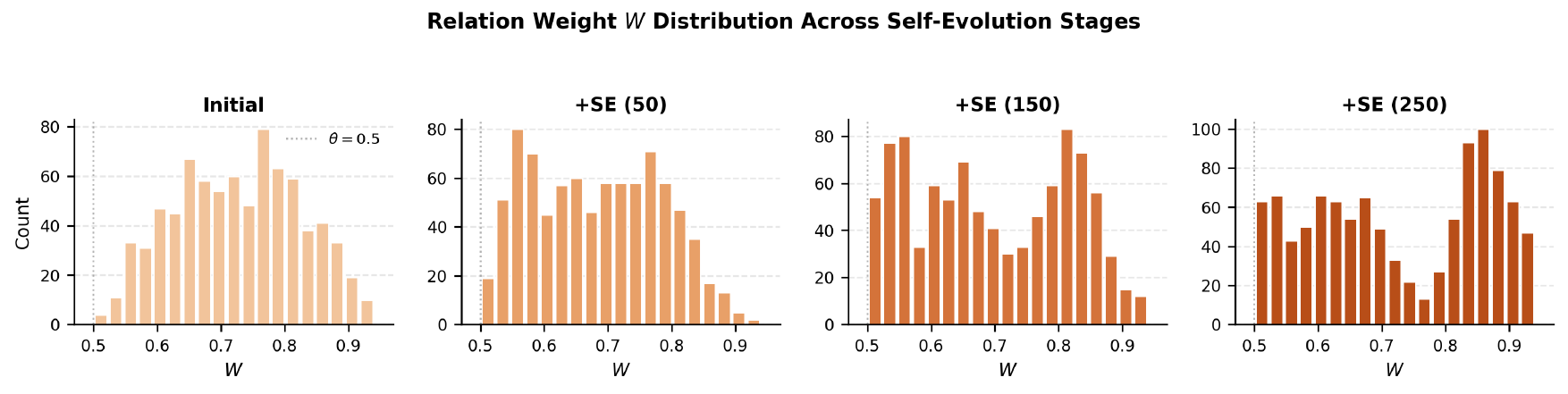}
    \caption{Distribution of relation weight $W$ in self-evolving stage.}
    \label{fig:W_cropped}
\end{figure*}

\paragraph{Mem0}
Mem0 \citep{chhikara2025mem0} is a long-term conversational memory architecture designed to maintain persistent user information across extended, multi-session interactions. It dynamically extracts facts from ongoing dialogue and stores them as concise natural language memories. When new information arrives, Mem0 compares extracted memories against existing ones and applies memory management operations such as addition, update, deletion, or no-op to maintain consistency and avoid redundancy. During inference, relevant memories are retrieved via embedding-based similarity search and provided to LLM as compact context, enabling efficient and coherent long-term reasoning without relying on full conversation history.

\paragraph{Mem0$^{g}$}
Mem0$^{g}$ extends the Mem0 framework by incorporating a graph-based memory representation to explicitly model entities and their relationships. Memories are stored as labeled nodes and directed edges, capturing structured relational information such as temporal order and semantic dependencies. Memory extraction converts dialogue into entity-relation triplets, which are incrementally integrated into a persistent knowledge graph with conflict detection and update mechanisms. At query time, Mem0$^{g}$ retrieves relevant subgraphs using a combination of entity-centric traversal and semantic triplet matching, providing structurally grounded context for answering queries that require multi-step or temporal reasoning.

\paragraph{A-Mem}
A-Mem \citep{xu2025mem} is an agentic memory system designed to support long-term knowledge organization for LLM-based agents. A-Mem stores interactions as atomic memory notes enriched with structured attributes such as contextual descriptions, keywords, tags, and dense embeddings. When new memories are added, the system autonomously establishes semantic links with existing memories and may further evolve historical notes by updating their contextual representations. Memory retrieval is performed via embedding-based similarity search, during which related memories connected through the agentic linkage structure are jointly accessed, enabling more coherent multi-hop reasoning and adaptive long-term memory usage.

\paragraph{ReMe}
ReMe \citep{cao2025remember} is a dynamic procedural memory framework for experience-driven evolution of LLM-based agents. Rather than storing raw interaction histories, ReMe distills reusable experiences from past trajectories, covering both successful patterns and failure-derived lessons. Retrieved experiences are adapted to new tasks via scenario-aware indexing, reranking, and rewriting. A utility-based refinement mechanism further maintains memory quality by incorporating high-value experiences and pruning ineffective ones, enabling long-term improvement without parameter updates.

\paragraph{FLEX}
FLEX \citep{cai2025flex} is a gradient-free learning paradigm that enables LLM agents to continuously evolve by accumulating and reusing experiential knowledge. Rather than updating model parameters, FLEX constructs an explicit experience library by distilling semantic insights from successful and failed interaction trajectories through forward exploration and reflection. The experience library is incrementally updated via a dedicated updater module and serves as an external knowledge substrate that guides future reasoning through selective retrieval. By decoupling learning from parameter optimization, FLEX supports scalable experience accumulation and enables experience inheritance across agents, allowing knowledge acquired by one agent to be transferred to others in a plug-and-play manner.

\section{Implementation Details}

\subsection{Hyperparameters}
\label{app:hyperparameters}

The four edge weight similarity coefficients $\alpha, \beta, \gamma, \delta$ are each set to $0.25$, applying uniform weighting across $S_{\text{entity}}$, $S_{\text{structure}}$, $S_{\text{synergy}}$, and $S_{\text{task}}$ to avoid introducing manual bias toward any single signal. During graph traversal, $K_n=5$ neighbor candidates are exposed per step with a maximum budget of $T_{\text{max}}=60$ steps. Semantic similarity scoring is performed using DeepSeek-V3.2.

\subsection{Similarity Calculation}

\label{app:similarity}
 
We describe the four similarity signals used to initialize edge weights in $\mathcal{G}^{exp}$.
 
\paragraph{Entity Similarity}
 
Each experience is parsed into normalized core entities spanning five roles (Condition, Constraint, Action, Rationale, Outcome). For each experience $e_i$, a TF-IDF weighted entity vector
$x_i \in \mathbb{R}^{|\mathcal{U}|}$ is constructed over the entity
vocabulary $\mathcal{U}$, where each entry is
$x_i(u) = \mathrm{tf}(u, e_i) \cdot \mathrm{idf}(u)$,
with $\mathrm{tf}(u, e_i)$ denoting the frequency of entity $u$
in $e_i$ and IDF defined as:
\begin{equation}
    \mathrm{idf}(u) = \log\frac{N+1}{df(u)+1} + 1
\end{equation}
where $N=|\mathcal{V}|$ is the total number of experiences and $df(u)$ is the number of experiences containing entity $u$. Entity similarity is the cosine similarity between vectors:
\begin{equation}
    S_{\text{entity}}(e_i, e_j) = \frac{x_i \cdot x_j}{\|x_i\|\,\|x_j\|}
\end{equation}
 
\paragraph{Structure Similarity}
 
Each experience is decomposed into role-edge paths of length $k \in \{1,2,3,4\}$ and paths are represented as sequences of role-typed edges, capturing the reasoning skeleton rather than surface content. Let $P_k(e_i)$ denote the set of all role-edge paths of length $k$
extracted from experience $e_i$. For each path length $k$, the Jaccard similarity over path sets is:
\begin{equation}
    S_k(e_i, e_j) = \frac{|P_k(e_i) \cap P_k(e_j)|}{|P_k(e_i) \cup P_k(e_j)|}
\end{equation}
The four levels are combined with weights proportional to path length, reflecting that longer paths encode more complete reasoning chains:
\begin{equation}
    S_{\text{structure}}(e_i, e_j) = \sum_{k=1}^{4} \frac{k}{\sum_{l=1}^{4} l}\, S_k(e_i, e_j)
\end{equation}

\begin{figure*}[t]
    \centering
    \includegraphics[width=\textwidth]{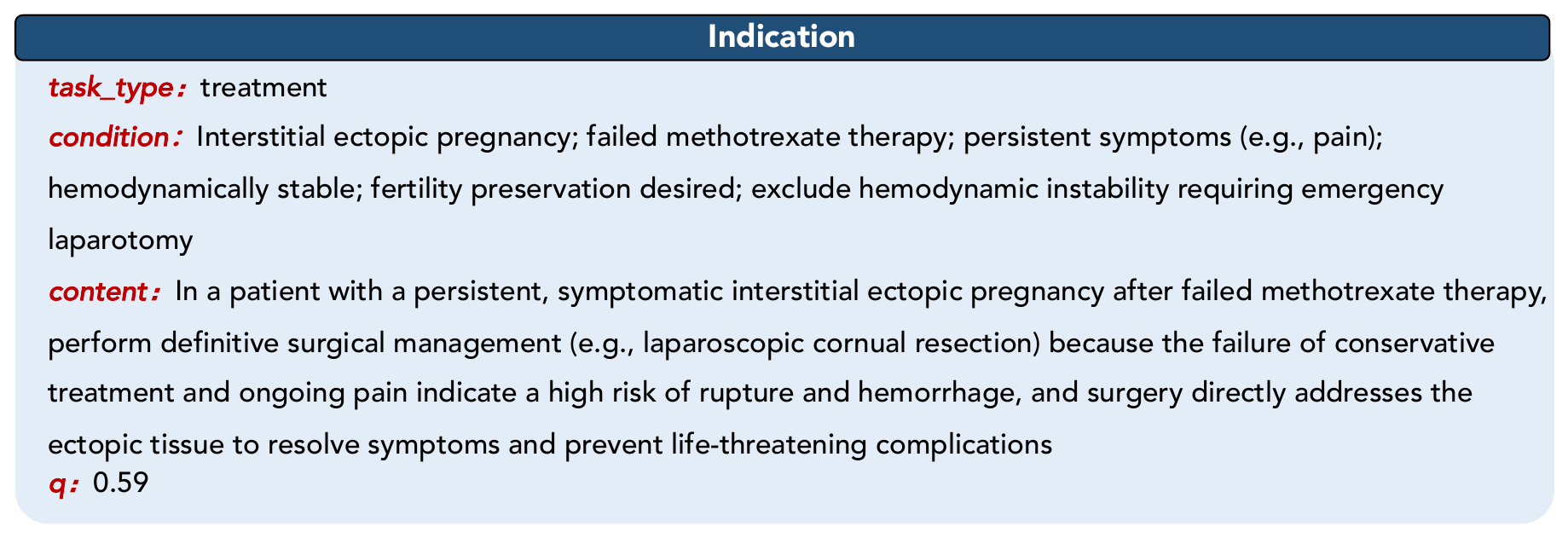}
    \caption{Examples of experiences, illustrating the four-field structure $(c_i, s_i, z_i, Q_i)$ defined in Section~\ref{sec:experience}.}
    \label{fig:experience_example}
\end{figure*}
 
\paragraph{Semantic Similarity}
 
Semantic similarity is computed via LLM-based scoring (Table~\ref{tab:prompt_sim_sys}), to capture joint-use value: whether two experiences are suitable to be co-retrieved and co-applied under the same query. For two experiences $e_i$ and $e_j$, the model produces
$S_{\text{synergy}}(e_i, e_j) \in [0, 1]$.
 
\paragraph{Task Similarity}
 
Task similarity is a binary indicator of task-type consistency:
\begin{equation}
    S_{\text{task}}(e_i, e_j) = \mathbf{1}[\tau_i = \tau_j]
\end{equation}
where $\tau_i$ and $\tau_j$ denote the task types of $e_i$ and $e_j$, automatically assigned by the LLM during experience extraction.

\section{Additional Experimental Results}
\label{app:reasoning_detail}
 
Table~\ref{tab:reasoning_detail_diag} and Table~\ref{tab:reasoning_detail_treat} reports per-dimension reasoning scores (Efficiency, Factuality, Completeness) and their average score (R) for each method on MedR-Bench diagnosis (n=329) and treatment (n=148) tasks respectively. All values are reported as mean$_{\pm\text{half-CI}}$ under the two-sided 95\% z-based confidence interval following the evaluation protocol of \citet{MedRBench}.

\section{Dual-layer Memory Graph}

We provide additional details of the proposed dual-layer memory graph, including concrete experience examples and visualizations of both the entity layer and the experience layer, to facilitate intuitive understanding.

\subsection{Visualization of the Entity-layer Graph}
\label{app:entity_graph}

\begin{figure}[t]
    \centering
    \includegraphics[width=\linewidth]{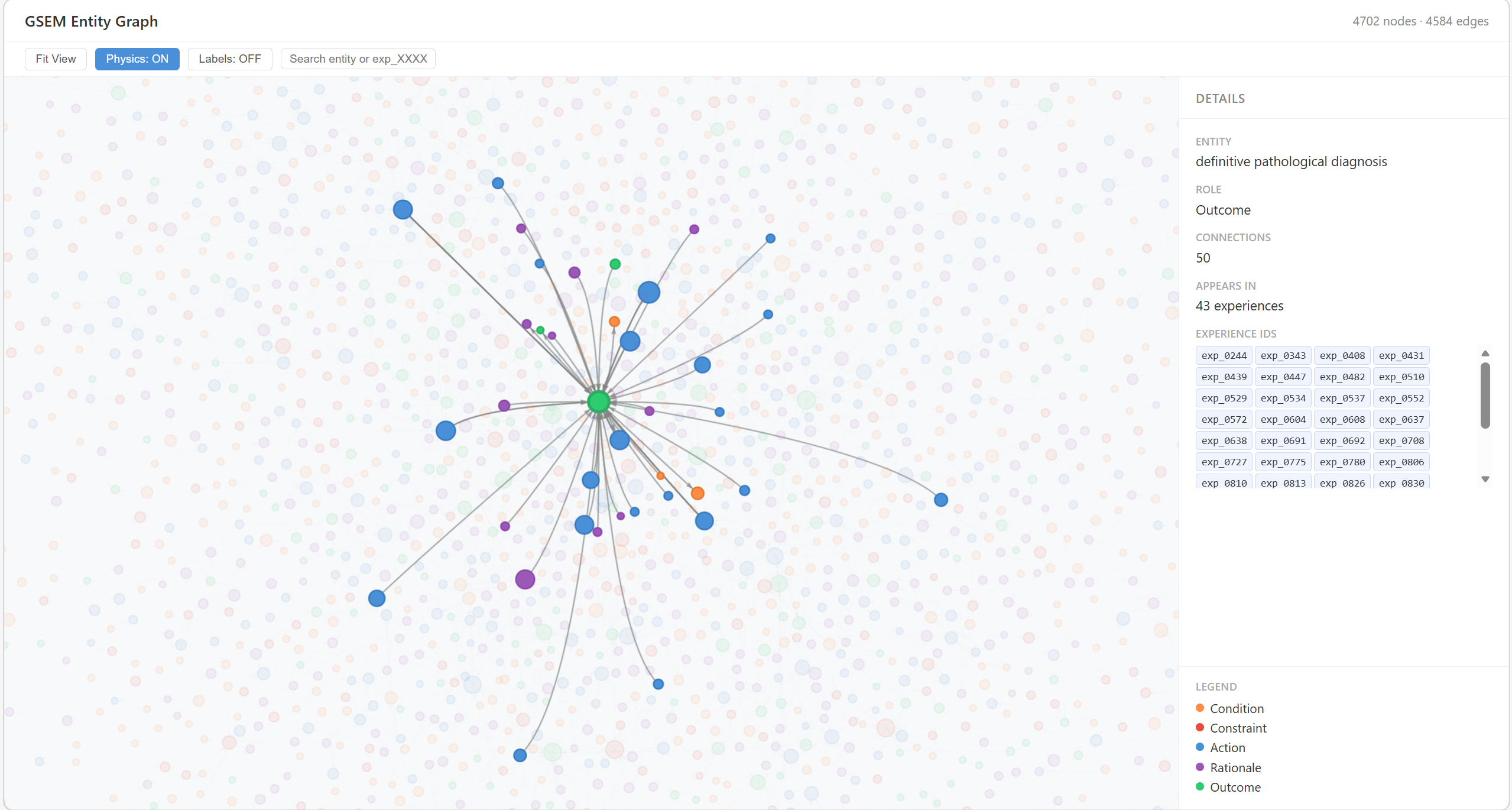}
    \caption{Visualization of a subgraph in the entity layer
    $\mathcal{G}^{ent}$. Nodes correspond to typed decision entities, and edges capture internal reasoning flows within and across experiences.}
    \label{fig:entity_graph}
\end{figure}

Figure~\ref{fig:entity_graph} visualizes a local subgraph of the entity layer $\mathcal{G}^{ent}$. Each node corresponds to a decision entity with a specific semantic role, including \textit{Condition}, \textit{Constraint}, \textit{Action}, \textit{Rationale}, and \textit{Outcome}. Node colors indicate entity types, and directed edges represent the internal reasoning flow within experiences. This entity-level structure enables fine-grained similarity estimation and supports interpretable experience retrieval.

\subsection{Visualization of the Experience-layer Graph}
\label{app:experience_graph}

\begin{figure}[t]
    \centering
    \includegraphics[width=\linewidth]{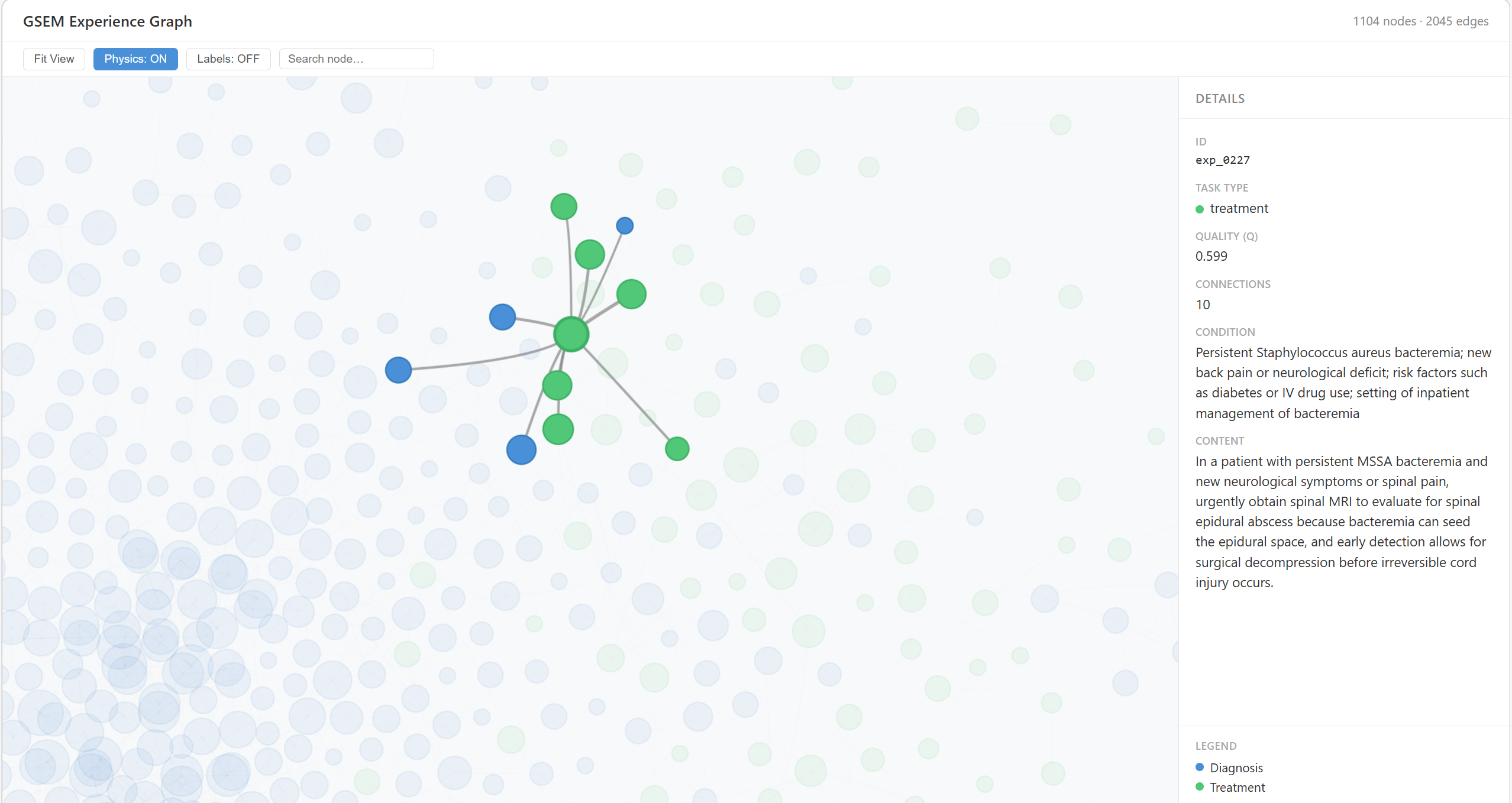}
    \caption{Visualization of a subgraph in the experience layer
    $\mathcal{G}^{exp}$. Node size encodes experience quality $Q_i$, and edge
    weights reflect learned inter-experience associations.}
    \label{fig:experience_graph}
\end{figure}

Figure~\ref{fig:experience_graph} illustrates a subgraph of the experience layer $\mathcal{G}^{exp}$. Each node represents an experience, and node size reflects its quality score $Q_i$. Directed edges represent inter-experience relations, with edge weights $W_{ij}$ capturing association strength derived from entity overlap, structural similarity, synergy (joint-use value) and task consistency.

\begin{table*}[h]
\centering
\begin{tabular}{lcccc}
\toprule
Method & Efficiency & Factuality & Completeness & Diagnosis/R (AVG) \\
\midrule
GraphRAG & $\mathbf{0.983}_{\pm.014}$ & $0.842_{\pm.039}$ & $0.782_{\pm.045}$ & $0.869_{\pm.036}$ \\
A-Mem    & $\mathbf{0.983}_{\pm.014}$ & $0.610_{\pm.053}$ & $0.770_{\pm.045}$ & $0.788_{\pm.044}$ \\
FLEX     & $\mathbf{0.983}_{\pm.014}$ & $0.588_{\pm.053}$ & $0.740_{\pm.047}$ & $0.770_{\pm.045}$ \\
\midrule
GSEM     & $0.978_{\pm.016}$ & $\mathbf{0.852}_{\pm.038}$ & $\mathbf{0.793}_{\pm.044}$ & $\mathbf{0.874}_{\pm.036}$ \\
\bottomrule
\end{tabular}
\caption{Per-dimension reasoning quality on MedR-Bench
Diagnosis (n=329) with 95\% CI.}
\label{tab:reasoning_detail_diag}
\end{table*}
 
\begin{table*}[h]
\centering
\begin{tabular}{lcccc}
\toprule
Method & Efficiency & Factuality & Completeness & Treatment/R (AVG) \\
\midrule
GraphRAG & $\mathbf{0.993}_{\pm.013}$ & $\mathbf{0.918}_{\pm.044}$ & $\mathbf{0.760}_{\pm.069}$ & $\mathbf{0.890}_{\pm.050}$ \\
A-Mem    & $0.992_{\pm.014}$ & $0.815_{\pm.063}$ & $0.709_{\pm.073}$ & $0.839_{\pm.059}$ \\
FLEX     & $0.990_{\pm.016}$ & $0.662_{\pm.076}$ & $0.667_{\pm.076}$ & $0.773_{\pm.067}$ \\
\midrule
GSEM     & $0.989_{\pm.017}$ & $0.836_{\pm.060}$ & $0.704_{\pm.074}$ & $0.843_{\pm.059}$ \\
\bottomrule
\end{tabular}
\caption{Per-dimension reasoning quality on MedR-Bench
Treatment (n=148) with 95\% CI.}
\label{tab:reasoning_detail_treat}
\end{table*}

\subsection{Experience Examples}
\label{app:examples}

Figure~\ref{fig:experience_example} shows an experience extracted from clinical reasoning trajectories, illustrating the structure of an atomic memory unit. Each experience $e_i=(c_i,s_i,z_i,Q_i)$ encodes reusable decision knowledge in a specific clinical context.

\begin{itemize}[noitemsep, topsep=0pt, leftmargin=*]
    \item \textbf{Condition} ($c_i$) specifies the clinical context under which the experience is applicable, summarizing the key situational patterns that trigger the decision.
    \item \textbf{Content} ($s_i$) describes the core decision strategy or action recommendation that is expected to be useful under the given condition.
    \item \textbf{Polarity} ($z_i \in \{\oplus,\ominus\}$) indicates whether the experience represents an \textit{Indication} (a strategy that has led to successful outcomes) or a \textit{Contraindication} (a pattern associated with failures and thus to be avoided).
    \item \textbf{Quality Score} ($Q_i\in[0,1]$) reflects the current reliability of the experience, initialized during memory construction and continuously refined through online task feedback.
\end{itemize}

\section{Case Study}
\label{app:case_study}

\paragraph{Case 1: Boundary Failure}
We present a representative case illustrating the \textbf{Boundary Failure} mode, where an inapplicable experience misleads reasoning by overriding a critical clinical boundary condition (Table~\ref{tab:case_study}). ReMe retrieves exp1 (conflicting data interpretation), whose applicability boundary requires a genuine diagnostic conflict. Here, spontaneous regression is not conflicting data but a meaningful clinical signal indicating active immune control. Applying exp1 causes ReMe to dismiss the abscopal effect and initiate unnecessary systemic therapy, violating the key boundary condition of ``watchful waiting during ongoing regression.'' GSEM, by contrast, correctly identifies the abscopal effect as an immune-mediated response and defers systemic therapy pending confirmation of regression.

\paragraph{Case 2: Collaboration Failure}
We present a representative case illustrating the \textbf{Collaboration Failure} mode, where co-retrieved experiences lack joint-use value and fail to provide coherent guidance for a critical surgical decision (Table~\ref{tab:case_study2}). ReMe retrieves exp1 (surgical steps for acute abdominal emergencies) and exp2 (treatment planning with pre-established diagnoses), which appear individually relevant but do not collaborate toward the key decision of whether to perform emergency pancreaticoduodenectomy (EPD) on-site: exp1 drives immediate step listing while exp2 accepts the injury list and defers pancreaticoduodenal reconstruction, together causing ReMe to miss the EPD decision entirely. GSEM, by contrast, retrieves three complementary experiences covering vascular hemorrhage control (exp\_1315), hollow viscus contamination management (exp\_0769), and hemodynamic resuscitation (exp\_0796), whose joint use coherently supports the conclusion that irreparable pancreatic head injury necessitates on-site EPD.

\begin{table*}[ht]
\centering
\caption{Case 1 (PMC11631726): Boundary Failure. A 50-year-old male with metastatic undifferentiated pleomorphic sarcoma (UPS) showing spontaneous regression attributed to the abscopal effect after multiple surgeries, radiotherapy, and two lines of chemotherapy. Ground truth: continue close follow-up leveraging the abscopal effect; do \emph{not} initiate systemic therapy while regression is ongoing.}
\label{tab:case_study}
\small
\begin{tabular}{p{2.8cm} p{6.2cm} p{6.2cm}}
\toprule
 & \textbf{GSEM} & \textbf{ReMe} \\
\midrule
\textbf{Retrieved Exps}
& exp\_0907: history of prior radiotherapy + new progressive deficit in irradiated field; exp\_0858: undifferentiated sarcoma with small round cell morphology, pre-treatment planning; + 6 other sarcoma-related experiences
& \textit{exp1}: interpreting conflicting clinical data (PET-negative lesions vs.\ known progressive disease); \textit{exp2}: mixed biomarker findings (driver mutation + PD-L1) in metastatic cancer \\
\midrule
\textbf{Key Steps}
& Step 2: Recognizes abscopal effect as immune-mediated response. Step 4: Decides close observation is appropriate while spontaneous regression is ongoing; defers systemic therapy unless progression occurs.
& Step 4: \textit{exp1} frames spontaneous regression as ``conflicting data'' and overrides it. Step 5--6: Proceeds to select third-line systemic therapy (pazopanib) despite ongoing regression. \\
\midrule
\textbf{Final Answer}
& Close surveillance with PET-CT in 6--8 weeks; initiate anti-PD1 immunotherapy \emph{only if} regression halts or progresses. \textbf{(Correct)}
& Initiate pazopanib (800 mg/day) immediately as third-line therapy; refer to immunotherapy clinical trial. \textbf{(Incorrect)} \\
\bottomrule
\end{tabular}
\end{table*}

\begin{table*}[ht]
\centering
\caption{Case 2 (PMC11490744): Collaboration Failure. A 49-year-old male with blunt abdominal trauma, hemodynamic instability, and multiple organ injuries including irreparable pancreatic head and duodenal injuries. Ground truth: perform emergency pancreaticoduodenectomy (Whipple procedure) combined with IVC repair, right nephrectomy, adrenalectomy, cholecystectomy, and liver hemostasis.}
\label{tab:case_study2}
\small
\begin{tabular}{p{2.8cm} p{6.2cm} p{6.2cm}}
\toprule
 & \textbf{GSEM} & \textbf{ReMe} \\
\midrule
\textbf{Retrieved Exps}
& exp\_1315: elderly male with hemodynamic collapse and retroperitoneal bleeding; exp\_0769: acute abdomen with pneumoperitoneum and hollow viscus perforation; exp\_0796: hospitalized patient with sudden severe abdominal pain and hemodynamic instability; + 3 other abdominal emergency experiences
& \textit{exp1}: generating surgical steps for acute abdominal emergencies; \textit{exp2}: formulating treatment plans when explicit diagnoses are provided; \textit{exp3}: dental trauma classification discrepancy (irrelevant); \textit{exp5}: IBD with pre-established diagnosis (irrelevant) \\
\midrule
\textbf{Key Reasoning Steps}
& Step 3: Identifies IVC laceration and gastroduodenal artery as immediate hemorrhage sources. Step 3--4: Recognizes pancreaticoduodenal injury severity and concludes EPD is required for irreparable injury; includes Whipple resection in the definitive surgical plan.
& Step 2: \textit{exp1} drives immediate surgical step listing without resolving EPD decision. Step 3: \textit{exp2} accepts injury list and moves to damage control; pancreaticoduodenal injury is deferred to staged reconstruction rather than addressed on-site. \\
\midrule
\textbf{Final Answer}
& Emergency damage control laparotomy including IVC repair, right nephrectomy, gastroduodenal artery ligation, and EPD (Whipple resection) for irreparable pancreatic head injury; postoperative ICU management. \textbf{(Correct)}
& Emergency exploratory laparotomy with IVC repair, nephrectomy, liver packing; pancreaticoduodenal injury deferred with drainage and staged reconstruction after 24--48 hours. \textbf{(Incorrect)} \\
\bottomrule
\end{tabular}
\end{table*}

\section{Prompt Templates}
\label{app:prompts}

\subsection{Indication Experience Extraction}

Tables~\ref{tab:prompt_indication_sys} and~\ref{tab:prompt_indication_human} show the system and human prompts used to extract Indication experiences from successful trajectories.

\begin{table*}[ht]
\centering
\begin{tcolorbox}[
  colback=white,
  colframe=black,
  boxrule=0.6pt,
  left=6pt,right=6pt,top=6pt,bottom=6pt,
  title={System Prompt~---~Indication Experience Extraction},
  coltitle=white,
  colbacktitle=black,
  fonttitle=\bfseries,
  width=\textwidth
]
\small
\setlength{\parindent}{0pt}
\setlength{\parskip}{0.4em}

You are a medical knowledge engineer extracting positive knowledge experiences (Indications) from successful clinical reasoning trajectories.

\textbf{DEFINITION:}\\
An Indication is a contextualized clinical reasoning experience that captures how medical knowledge is applied in practice to achieve successful diagnostic or therapeutic outcomes.

\textbf{ANALYSIS FRAMEWORK:}
\begin{itemize}[leftmargin=*, itemsep=2pt]
  \item \textbf{CLINICAL CONTEXT:} Identify patient characteristics, symptoms, and situational factors that define applicability.
  \item \textbf{ACTIONABLE GUIDANCE:} Specify the diagnostic or therapeutic approach taken, and highlight the decisive next step.
  \item \textbf{REASONING LOGIC:} Explain the causal relationships and rationale that connect context to action, focusing on the key cue or mechanism that makes the action appropriate.
  \item \textbf{OUTCOME BENEFIT:} Describe what this action improves in reasoning quality, such as avoiding misdiagnosis, reducing delay, or prioritizing a reversible cause.
\end{itemize}

\textbf{EXTRACTION PRINCIPLES:}
\begin{itemize}[leftmargin=*, itemsep=2pt]
  \item Focus on reasoning patterns embedded in successful trajectories.
  \item Capture the connection between clinical presentation, decision-making, and outcomes.
  \item Extract transferable knowledge applicable to similar clinical scenarios.
  \item Prefer specific decision-relevant guidance over generic textbook advice.
  \item Make applicability conditions explicit as short phrases that can serve as retrieval cues.
  \item Ground the experience in the provided trajectory and cite where the pattern appears.
  \item Use standard clinical and biomedical terminology. Prefer guideline-style terms, established disease names, and test names. Avoid informal wording and ambiguous lay descriptions.
\end{itemize}

\textbf{OUTPUT FORMAT:}\\
Generate 1--2 knowledge experiences as a valid JSON array. The output must be ONLY the JSON array with no additional text, explanation, or markdown code blocks.

\begin{verbatim}
[
  {
    "content":   "A coherent 2-3 sentence clinical reasoning experience
                  that states the context, action, rationale, benefit.",
    "condition": "Precise applicability conditions as short semicolon-
                  separated phrases.",
    "task_type": "Clinical task category as a short phrase.",
    "evidence":  "Trajectory references using trajectory IDs and steps."
  }
]
\end{verbatim}

\textit{CRITICAL: Output ONLY valid JSON array. No markdown, no explanation, no code blocks. Start with \texttt{[} and end with \texttt{]}.}

The experience should read as a clinical decision rule: \textit{``In [context], perform [action] because [rationale].''}
\end{tcolorbox}
\caption{System prompt for Indication experience extraction.}
\label{tab:prompt_indication_sys}
\end{table*}

\begin{table*}[ht]
\centering
\begin{tcolorbox}[
  colback=white,
  colframe=black,
  boxrule=0.6pt,
  left=6pt,right=6pt,top=6pt,bottom=6pt,
  title={Human Prompt~---~Indication Experience Extraction},
  coltitle=white,
  colbacktitle=black,
  fonttitle=\bfseries,
  width=\textwidth
]
\small
\setlength{\parindent}{0pt}
\setlength{\parskip}{0.4em}
\texttt{\# Case Information}\\
\texttt{\{case\_info\}}\\[6pt]
\texttt{\# Successful Reasoning Trajectories}\\
\texttt{\{trajectory\}}\\[6pt]
\texttt{Extract 1-2 Indication experiences as JSON array.}
\end{tcolorbox}
\caption{Human prompt for Indication experience extraction.}
\label{tab:prompt_indication_human}
\end{table*}

\subsection{Contraindication Experience Extraction --- Divergence Analysis}

Tables~\ref{tab:prompt_contra_sys} and~\ref{tab:prompt_contra_human} show the prompts used to identify fatal divergence points between successful and failed trajectories.

\begin{table*}[ht]
\centering
\begin{tcolorbox}[
  colback=white,
  colframe=black,
  boxrule=0.6pt,
  left=6pt,right=6pt,top=6pt,bottom=6pt,
  title={System Prompt~---~Divergence Analysis},
  coltitle=white,
  colbacktitle=black,
  fonttitle=\bfseries,
  width=\textwidth
]
\small
\setlength{\parindent}{0pt}
\setlength{\parskip}{0.4em}

You are a clinical reasoning analyst identifying fatal divergence points between successful and failed diagnostic trajectories.

\textbf{DEFINITION:}\\
A fatal divergence point is the earliest critical decision difference where the failed trajectory commits to an irreversible reasoning path that prevents correct diagnosis or treatment.

\textbf{DIVERGENCE ANALYSIS FRAMEWORK:}
\begin{itemize}[leftmargin=*, itemsep=2pt]
  \item \textbf{DECISION COMPARISON:} Compare reasoning decisions at each step between failure and success trajectories.
  \item \textbf{CRITICALITY ASSESSMENT:} Identify which decision difference directly caused the ultimate failure.
  \item \textbf{IRREVERSIBILITY ANALYSIS:} Determine the point where the failed trajectory cannot self-correct.
  \item \textbf{CONSEQUENCE TRACING:} Map how the divergent decision leads to the adverse outcome.
\end{itemize}

\textbf{IDENTIFICATION PRINCIPLES:}
\begin{itemize}[leftmargin=*, itemsep=2pt]
  \item Focus on the EARLIEST decision point where paths meaningfully separate.
  \item Identify decisions that are IRREVERSIBLE given subsequent trajectory constraints.
  \item Analyze CAUSAL LINKS between the divergent decision and final failure.
  \item Consider MEDICAL URGENCY and time-sensitive intervention windows.
\end{itemize}

\textbf{OUTPUT FORMAT:}\\
Generate divergence analysis as JSON object:

\begin{verbatim}
{
  "divergence_step": <integer>,
  "success_decision": "Decision in successful trajectory at this step",
  "failure_decision": "Erroneous decision in failed trajectory",
  "why_fatal": "Why this difference is fatal and irreversible",
  "consequence": "Adverse outcome pathway this divergence initiates"
}
\end{verbatim}

The analysis should capture: \textit{At step [N], success chose [A] while failure chose [B], which is fatal because [mechanism], leading to [consequence].}
\end{tcolorbox}
\caption{System prompt for Contraindication divergence analysis.}
\label{tab:prompt_contra_sys}
\end{table*}

\begin{table*}[ht]
\centering
\begin{tcolorbox}[
  colback=white,
  colframe=black,
  boxrule=0.6pt,
  left=6pt,right=6pt,top=6pt,bottom=6pt,
  title={Human Prompt~---~Divergence Analysis},
  coltitle=white,
  colbacktitle=black,
  fonttitle=\bfseries,
  width=\textwidth
]
\small
\setlength{\parindent}{0pt}
\setlength{\parskip}{0.4em}
\texttt{\# Successful Trajectory}\\
\texttt{\{success\_trajectory\}}\\[6pt]
\texttt{\# Failed Trajectory}\\
\texttt{\{failure\_trajectory\}}\\[6pt]
\texttt{\# Task Goal}\\
\texttt{The correct outcome is: \{gold\_answer\}}\\
\texttt{The failed trajectory resulted in: \{wrong\_answer\}}\\[6pt]
\texttt{Identify the fatal divergence point as JSON object.}
\end{tcolorbox}
\caption{Human prompt for Contraindication divergence analysis.}
\label{tab:prompt_contra_human}
\end{table*}

\subsection{Contraindication Experience Extraction}

Tables~\ref{tab:prompt_contra_extraction_sys} and~\ref{tab:prompt_contra_extraction_human} show the prompts used to extract Contraindication experiences from the divergence analysis.

\begin{table*}[ht]
\centering
\begin{tcolorbox}[
  colback=white,
  colframe=black,
  boxrule=0.6pt,
  left=6pt,right=6pt,top=6pt,bottom=6pt,
  title={System Prompt~---~Contraindication Experience Extraction},
  coltitle=white,
  colbacktitle=black,
  fonttitle=\bfseries,
  width=\textwidth
]
\small
\setlength{\parindent}{0pt}
\setlength{\parskip}{0.4em}

You are a clinical safety analyst extracting contraindication experiences from failed reasoning trajectories.

\textbf{DEFINITION:}\\
A Contraindication is a reasoning error pattern learned from diagnostic or therapeutic failures, capturing what approaches should be avoided in specific clinical contexts to prevent adverse outcomes.

\textbf{ANALYSIS FRAMEWORK:}
\begin{itemize}[leftmargin=*, itemsep=2pt]
  \item \textbf{ERROR PATTERN:} Identify the specific reasoning or decision-making error that occurred.
  \item \textbf{CLINICAL CONTEXT:} Specify the situation where this error led to failure.
  \item \textbf{FAILURE MECHANISM:} Analyze how this error caused diagnostic delay or therapeutic harm through a clear causal chain.
  \item \textbf{CONSEQUENCE ANALYSIS:} Document the failure outcome that resulted, such as a missed critical step, wrong decision, or delayed diagnosis.
\end{itemize}

\textbf{EXTRACTION PRINCIPLES:}
\begin{itemize}[leftmargin=*, itemsep=2pt]
  \item Focus on systematic and preventable errors in the reasoning process rather than random mistakes.
  \item Prefer concrete decision errors over abstract cognitive-bias labels.
  \item Use the provided failure-success divergence analysis to anchor the contraindication in the concrete difference between failure and success.
  \item Frame the insight as a safety guideline with explicit risk scenarios and clear boundaries.
  \item Ground the contraindication in the provided trajectories and cite where the failure emerges.
  \item Use standard clinical and biomedical terminology. Prefer established disease names, test names, and treatment terms. Avoid informal wording and ambiguous lay descriptions.
\end{itemize}

\textbf{OUTPUT FORMAT:}\\
Generate 1 contraindication experience as JSON object:

\begin{verbatim}
{
  "content":   "A coherent 2-3 sentence safety experience that states
                the context, the prohibited error pattern, the failure
                mechanism, and the consequence.",
  "condition": "Specific risk scenario as short semicolon-separated
                phrases. Include population, presentation, high-risk
                cues, and timing or setting when relevant.",
  "task_type": "Clinical task category as a short phrase.",
  "evidence":  "Failure-success references using case or trajectory IDs,
                divergence point, and step numbers."
}
\end{verbatim}

The experience should read as a safety warning: \textit{``In [context], do not [error] because [mechanism] leads to [harm].''}
\end{tcolorbox}
\caption{System prompt for Contraindication experience extraction.}
\label{tab:prompt_contra_extraction_sys}
\end{table*}

\begin{table*}[ht]
\centering
\begin{tcolorbox}[
  colback=white,
  colframe=black,
  boxrule=0.6pt,
  left=6pt,right=6pt,top=6pt,bottom=6pt,
  title={Human Prompt~---~Contraindication Experience Extraction},
  coltitle=white,
  colbacktitle=black,
  fonttitle=\bfseries,
  width=\textwidth
]
\small
\setlength{\parindent}{0pt}
\setlength{\parskip}{0.4em}
\texttt{\# Case Information}\\
\texttt{\{case\_info\}}\\[6pt]
\texttt{\# Reference Analysis (for context)}\\
\texttt{\{reference\_analysis\}}\\[6pt]
\texttt{\# Failure-Success Divergence Analysis}\\
\texttt{\{divergence\}}\\[6pt]
\texttt{\# Failure Trajectory (Original)}\\
\texttt{\{failure\_trajectory\}}\\[6pt]
\texttt{\# Success Trajectory (Original)}\\
\texttt{\{success\_trajectory\}}\\[6pt]
\texttt{Extract 1 Contraindication experience as JSON object.}
\end{tcolorbox}
\caption{Human prompt for Contraindication experience extraction.}
\label{tab:prompt_contra_extraction_human}
\end{table*}

\subsection{Entity Extraction}

Tables~\ref{tab:prompt_entity_sys} and~\ref{tab:prompt_entity_human} show the prompts used to extract decision-structure entities from each experience.

\begin{table*}[ht]
\centering
\begin{tcolorbox}[
  colback=white,
  colframe=black,
  boxrule=0.6pt,
  left=6pt,right=6pt,top=6pt,bottom=6pt,
  title={System Prompt~---~Entity Extraction},
  coltitle=white,
  colbacktitle=black,
  fonttitle=\bfseries,
  width=\textwidth
]
\small
\setlength{\parindent}{0pt}
\setlength{\parskip}{0.4em}

You are extracting a compact set of medical decision-structure entities from a reusable clinical reasoning experience.

\textbf{GOAL:}\\
Extract only the most decision-driving medical entities that define the clinical situation and the actionable strategy, while keeping the entity list small and easy to normalize.

\textbf{ROLE SCHEMA (fixed):}
\begin{itemize}[leftmargin=*, itemsep=2pt]
  \item \textbf{Condition:} Diagnoses, symptoms, findings, clinical states, patient status. Keep 1--2 anchors that determine whether the strategy applies.
  \item \textbf{Constraint:} Contraindications, feasibility limits, special risks. Keep at most 1 anchor; if none, omit.
  \item \textbf{Action:} Diagnostic tests, treatments, interventions, procedures. Keep 1--2 anchors that truly change the decision.
  \item \textbf{Rationale:} Medical reasoning basis, mechanism, justification. Keep at most 1 anchor; if none, omit.
  \item \textbf{Outcome:} Intended clinical goal or effect. Keep at most 1 anchor; if none, omit.
\end{itemize}

\textbf{EXTRACTION RULES:}
\begin{itemize}[leftmargin=*, itemsep=2pt]
  \item Extract only medically meaningful, decision-driving anchor entities.
  \item Use standard medical terminology; avoid generic words.
  \item \textbf{CRITICAL:} Each entity MUST be a noun or noun phrase of EXACTLY 1--3 words.
  \item Do not duplicate the same meaning across roles.
  \item Keep the total set compact (typically 5--8 entities).
\end{itemize}

\textbf{OUTPUT FORMAT:}
\begin{verbatim}
{
  "core_entities": [
    {"entity": "...", "role": "Condition"},
    {"entity": "...", "role": "Action"}
  ]
}
\end{verbatim}
\end{tcolorbox}
\caption{System prompt for entity extraction.}
\label{tab:prompt_entity_sys}
\end{table*}

\begin{table*}[ht]
\centering
\begin{tcolorbox}[
  colback=white,
  colframe=black,
  boxrule=0.6pt,
  left=6pt,right=6pt,top=6pt,bottom=6pt,
  title={Human Prompt~---~Entity Extraction},
  coltitle=white,
  colbacktitle=black,
  fonttitle=\bfseries,
  width=\textwidth
]
\small
\setlength{\parindent}{0pt}
\setlength{\parskip}{0.4em}
\texttt{\#\# Condition}\\
\texttt{\{condition\}}\\[6pt]
\texttt{\#\# Content}\\
\texttt{\{content\}}\\[6pt]
\texttt{Extract the core decision-structure entities and roles as specified.}\\
\texttt{Return a valid JSON object with field `core\_entities'.}\\
\texttt{Output only the JSON, no other text.}
\end{tcolorbox}
\caption{Human prompt for entity extraction.}
\label{tab:prompt_entity_human}
\end{table*}

\subsection{Role-Edge Structure Extraction}

Tables~\ref{tab:prompt_edge_sys} and~\ref{tab:prompt_edge_human} show the prompts used to extract the role-edge reasoning skeleton from each experience.

\begin{table*}[ht]
\centering
\begin{tcolorbox}[
  colback=white,
  colframe=black,
  boxrule=0.6pt,
  left=6pt,right=6pt,top=6pt,bottom=6pt,
  title={System Prompt~---~Role-Edge Structure Extraction},
  coltitle=white,
  colbacktitle=black,
  fonttitle=\bfseries,
  width=\textwidth
]
\small
\setlength{\parindent}{0pt}
\setlength{\parskip}{0.4em}

You are extracting the decision-flow structure (reasoning skeleton) of a clinical experience using role-to-role edges.

\textbf{ALLOWED ROLE-EDGES:}
\begin{itemize}[leftmargin=*, itemsep=1pt]
  \item Condition$\rightarrow$Action, Condition$\rightarrow$Condition, Condition$\rightarrow$Constraint, Condition$\rightarrow$Outcome, Condition$\rightarrow$Rationale
  \item Constraint$\rightarrow$Action, Constraint$\rightarrow$Rationale, Constraint$\rightarrow$Outcome
  \item Action$\rightarrow$Outcome, Action$\rightarrow$Rationale, Action$\rightarrow$Constraint, Action$\rightarrow$Action
  \item Rationale$\rightarrow$Action, Rationale$\rightarrow$Outcome, Rationale$\rightarrow$Constraint
\end{itemize}

\textbf{EXTRACTION RULES:}
\begin{itemize}[leftmargin=*, itemsep=2pt]
  \item Only use edges from the allowed list above.
  \item Include an edge only if it reflects a clear reasoning transition in the text.
  \item Every role-edge MUST be grounded by at least one entity-edge. Do NOT output a role-edge without a concrete entity pair.
  \item Prefer a minimal but expressive skeleton (typically 2--6 edges).
\end{itemize}

\textbf{OUTPUT FORMAT:}
\begin{verbatim}
{
  "role_edges": ["Condition->Constraint", "Constraint->Action", ...],
  "entity_edges": [
    {
      "edge": "Condition->Constraint",
      "from_entity": "<entity from core_entities>",
      "to_entity":   "<entity from core_entities>"
    }
  ]
}
\end{verbatim}
\end{tcolorbox}
\caption{System prompt for role-edge structure extraction.}
\label{tab:prompt_edge_sys}
\end{table*}

\begin{table*}[ht]
\centering
\begin{tcolorbox}[
  colback=white,
  colframe=black,
  boxrule=0.6pt,
  left=6pt,right=6pt,top=6pt,bottom=6pt,
  title={Human Prompt~---~Role-Edge Structure Extraction},
  coltitle=white,
  colbacktitle=black,
  fonttitle=\bfseries,
  width=\textwidth
]
\small
\setlength{\parindent}{0pt}
\setlength{\parskip}{0.4em}
\texttt{\#\# Core Entities with Roles}\\
\texttt{\{core\_entities\_json\}}\\[6pt]
\texttt{\#\# Condition}\\
\texttt{\{condition\}}\\[6pt]
\texttt{\#\# Content}\\
\texttt{\{content\}}\\[6pt]
\texttt{Extract the role-edge decision-flow structure and ground each}\\
\texttt{role-edge to specific entity pairs from the core entities.}\\
\texttt{Output only the JSON, no other text.}
\end{tcolorbox}
\caption{Human prompt for role-edge structure extraction.}
\label{tab:prompt_edge_human}
\end{table*}

\subsection{Semantic Similarity Evaluation}

Tables~\ref{tab:prompt_sim_sys} and~\ref{tab:prompt_sim_human} show the prompts used to evaluate semantic similarity between experience pairs for edge weight computation.

\begin{table*}[ht]
\centering
\begin{tcolorbox}[
  colback=white,
  colframe=black,
  boxrule=0.6pt,
  left=6pt,right=6pt,top=6pt,bottom=6pt,
  title={System Prompt~---~Semantic Similarity Evaluation},
  coltitle=white,
  colbacktitle=black,
  fonttitle=\bfseries,
  width=\textwidth
]
\small
\setlength{\parindent}{0pt}
\setlength{\parskip}{0.4em}

You are evaluating the semantic similarity between two clinical reasoning experiences.

\textbf{EVALUATION CRITERIA:}
\begin{itemize}[leftmargin=*, itemsep=2pt]
  \item \textbf{Clinical context similarity:} Do they address similar medical situations, conditions, or patient populations?
  \item \textbf{Strategy similarity:} Do they use similar diagnostic or treatment approaches?
  \item \textbf{Reasoning logic:} Do they follow similar decision-making patterns?
  \item \textbf{Transferability:} Would insights from one be applicable to the other?
\end{itemize}

\textbf{SCORING SCALE:}
\begin{itemize}[leftmargin=*, itemsep=2pt]
  \item 0.0: Completely unrelated (different domains, conditions, or strategies)
  \item 0.3: Weakly related (share some general medical principles but different contexts)
  \item 0.5: Moderately related (share some clinical aspects or reasoning patterns)
  \item 0.7: Highly related (similar conditions, strategies, or decision patterns)
  \item 1.0: Nearly identical (same condition and strategy with minor variations)
\end{itemize}

\textbf{OUTPUT FORMAT:}
\begin{verbatim}
{
  "similarity": <float between 0.0 and 1.0>,
  "reason": "<brief explanation in 1-2 sentences>"
}
\end{verbatim}
\end{tcolorbox}
\caption{System prompt for semantic similarity evaluation.}
\label{tab:prompt_sim_sys}
\end{table*}

\begin{table*}[ht]
\centering
\begin{tcolorbox}[
  colback=white,
  colframe=black,
  boxrule=0.6pt,
  left=6pt,right=6pt,top=6pt,bottom=6pt,
  title={Human Prompt~---~Semantic Similarity Evaluation},
  coltitle=white,
  colbacktitle=black,
  fonttitle=\bfseries,
  width=\textwidth
]
\small
\setlength{\parindent}{0pt}
\setlength{\parskip}{0.4em}
\texttt{\#\# Experience A}\\
\texttt{Condition: \{condition\_a\}}\\
\texttt{Content: \{content\_a\}}\\[6pt]
\texttt{\#\# Experience B}\\
\texttt{Condition: \{condition\_b\}}\\
\texttt{Content: \{content\_b\}}\\[6pt]
\texttt{Evaluate the semantic similarity between these two experiences.}\\
\texttt{Return a valid JSON object with `similarity' (float) and `reason' (string).}\\
\texttt{Output only the JSON, no other text.}
\end{tcolorbox}
\caption{Human prompt for semantic similarity evaluation.}
\label{tab:prompt_sim_human}
\end{table*}
\end{document}